\documentclass[10pt,twocolumn,letterpaper]{article}

\usepackage[pagenumbers]{cvpr} 

\usepackage{graphicx}
\usepackage{amsmath}
\usepackage{amssymb}
\usepackage{booktabs}

\usepackage{algpseudocode}
\usepackage{algorithm}
\usepackage{array}
\usepackage{multirow}
\usepackage{dsfont}
\usepackage[font={small}]{caption}
\usepackage{wrapfig,lipsum,booktabs}

\usepackage{color}
\usepackage{xcolor}
\definecolor{citecolor}{HTML}{0071bc}
\usepackage{xspace}

\usepackage[pagebackref=true,breaklinks=true,colorlinks,citecolor=citecolor,bookmarks=false]{hyperref}

\usepackage[capitalize]{cleveref}
\crefname{section}{Sec.}{Secs.}
\Crefname{section}{Section}{Sections}
\Crefname{table}{Table}{Tables}
\crefname{table}{Tab.}{Tabs.}

\newlength\savewidth\newcommand\shline{\noalign{\global\savewidth\arrayrulewidth
  \global\arrayrulewidth 1pt}\hline\noalign{\global\arrayrulewidth\savewidth}}

\newcommand{\IGNORE}[1]{}

\newcommand{\customfootnotetext}[2]{{
  \renewcommand{\thefootnote}{#1}
  \footnotetext[0]{#2}}}

\begin{document}

\title{CoordGAN: Self-Supervised Dense Correspondences Emerge from GANs}

\author{Jiteng Mu\textsuperscript{1*}, \quad
Shalini De Mello\textsuperscript{2}, \quad
Zhiding Yu\textsuperscript{2}, \quad
Nuno Vasconcelos\textsuperscript{1}, \quad \\
Xiaolong Wang\textsuperscript{1}, \quad
Jan Kautz\textsuperscript{2}, \quad
Sifei Liu\textsuperscript{2} \\
\textsuperscript{1}UC San Diego, \textsuperscript{2}Nvidia
}

\twocolumn[{
\vspace{-1em}
\maketitle
\vspace{-1em}

\begin{center}
    \centering
    \vspace{-0.2in}
    \includegraphics[width=\linewidth]{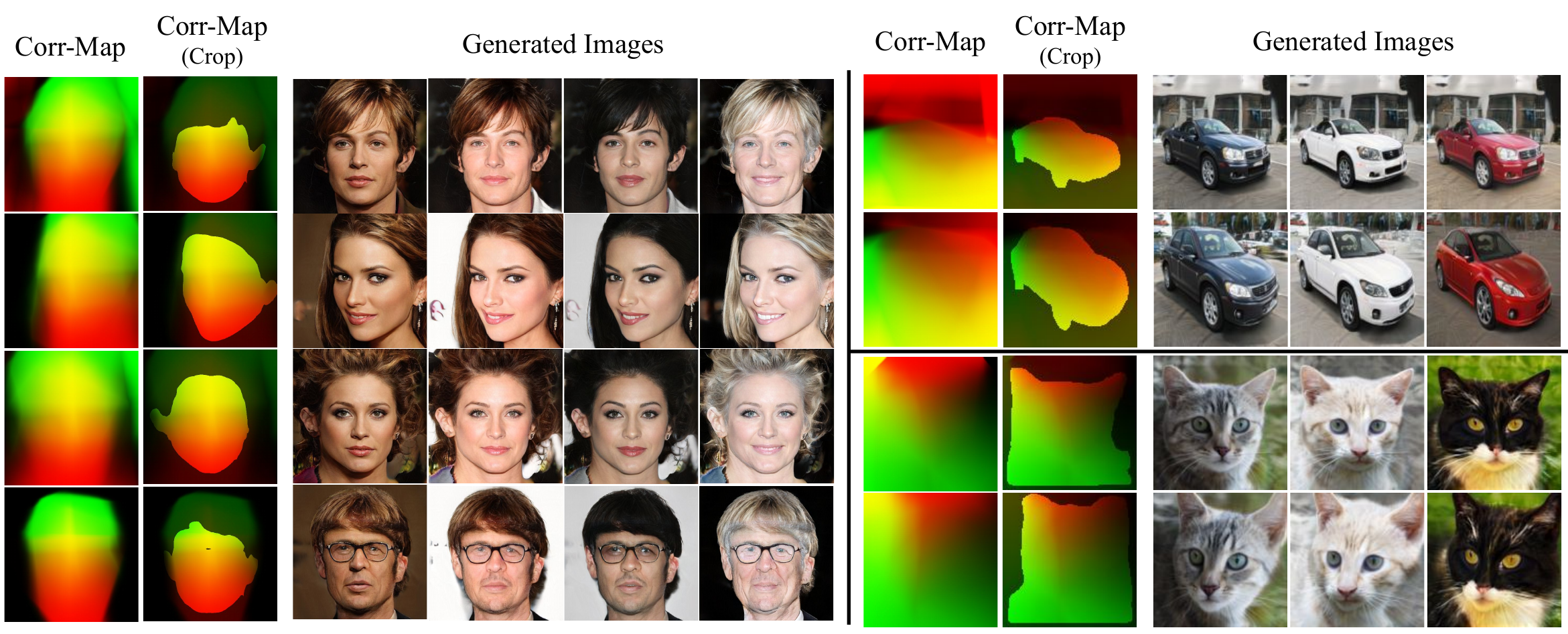}
    \vspace{-0.1in}
    \captionof{figure}{
    Images synthesized by the proposed CoordGAN for various object categories (left: faces; top-right: cars; bottom-right: cats) : each row displays images with the same structure but different textures; in each column, structure varies while keeping texture fixed. The correspondence maps (Corr-Map) controlling the structure of the synthesized images are shown in the first column of each row. For better visualization, we use off-the-shelf segmentation models to highlight the foreground areas of all the correspondence maps, as shown with Corr-Map (Crop).
    }
    \label{fig:teasing}
\end{center}
}]

\customfootnotetext{*}{Work done while an intern at Nvidia.}

\begin{abstract}
\vspace{-0.2in}

Recent advances show that Generative Adversarial Networks (GANs) can synthesize images with smooth variations along semantically meaningful latent directions, such as pose, expression, layout, etc. While this indicates that GANs implicitly learn pixel-level correspondences across images, few studies explored how to extract them explicitly. In this work, we introduce Coordinate GAN (CoordGAN), a structure-texture disentangled GAN that learns a dense correspondence map for each generated image. 
We represent the correspondence maps of different images as warped coordinate frames transformed from a canonical coordinate frame,
i.e., the correspondence map, which describes the structure (e.g., the shape of a face), is controlled via a transformation.
%
Hence, finding correspondences boils down to locating the same coordinate in different correspondence maps.
In CoordGAN, we sample a transformation to represent the structure of a synthesized instance, while an independent texture branch is responsible for rendering appearance details orthogonal to the structure. 
Our approach can also extract dense correspondence maps for real images by adding an encoder on top of the generator.
We quantitatively demonstrate the quality of the learned dense correspondences through segmentation mask transfer on multiple datasets. We also show that the proposed generator achieves better structure and texture disentanglement compared to existing approaches. Project page: \href{https://jitengmu.github.io/CoordGAN/}{https://jitengmu.github.io/CoordGAN/}
\end{abstract}


\vspace{-4mm}
\section{Introduction}
\label{sec:intro}
Generative Adversarial Networks (GANs) have  achieved great success in synthesizing high-quality images~\cite{radford2015unsupervised,karras2017progressive,brock2018large,karras2019style,karras2020analyzing}, and many recent studies show that they also learn a rich set of interpretable directions in the latent space~\cite{shen21closed,shen2020interpreting}. Moving latent codes along a semantically meaningful direction (e.g., pose) generates instances with smoothly varying appearance (e.g., continually changing viewpoints), implying that GANs also implicitly learn which pixels or regions are in correspondence with each other, from different synthesized instances. 

On the other hand, dense correspondence is established between local semantically-similar regions, but with varying appearance (e.g., patches of two different eyes). Learning dense correspondence across images of one category remains challenging because labeling large-scale, pixel-level annotations is extremely laborious. While most existing works rely on supervised~\cite{choy2016universal,rocco2017convolutional,han2017scnet,jeon2018parn}, or unsupervised~\cite{xiao2021learning} image classification networks, few have investigated how to learn dense correspondence from GANs.

In this work, we explore learning dense correspondence from GANs. Specifically, we aim to learn an explicit correspondence map, i.e., a pixel-level semantic label map. Since correspondence represents structure (e.g., shapes of facial components) and is independent of texture (e.g., global appearance like skin tone and texture), this task is highly relevant to disentanglement of structure and texture in GANs~\cite{wang16generative,nguyen-phuoc19hologan,yang21semantic,shen21closed,alharbiW20disentangled,kwon21diagonal}. Studies show that disentanglement of semantic attributes can be achieved by carefully searching for latent directions learned by GANs~\cite{harkonen2020ganspace,shen21closed,yang21semantic}, but all attributes being factorized have to be identified by humans. Some recent advances~\cite{alharbiW20disentangled,kwon21diagonal} demonstrate effective structure-texture disentanglement by improving the noise code input to GANs~\cite{alharbiW20disentangled}, or by applying spatial attention in the intermediate layers~\cite{kwon21diagonal}. However, they either produce a relatively low resolution (e.g., $4\times 4$) structure map~\cite{alharbiW20disentangled}, or do not produce it explicitly~\cite{kwon21diagonal}.

Our key idea is to introduce a novel coordinate space, from which pixel-level correspondence can be explicitly obtained for all the synthesised images of a category. Inspired by UV maps of 3D meshes~\cite{kanazawa2018learning,kulkarni2019canonical,li2020self}, where shapes of one category are represented as deformations of one canonical template, in this work, we represent the dense correspondence map of a generated image as a warped coordinate frame transformed from a canonical 2D coordinate map.
This enables the representation of a unique structure as a transformation between the warped and the canonical frames.
We design a Coordinate GAN (CoordGAN) with structure and texture controlled via two independently sampled noise vectors. While the texture branch controls the global appearance via Adaptive Instance Normalization (AdaIN)~\cite{karras2019style}, in the structure branch, we learn an MLP as the aforementioned transformation. This maps a sampled noise vector to a warped coordinate frame, which is further modulated in the generator to control the structure of the synthesized image in a hierarchical manner.

We adopt several objectives during training to ensure that the network learns accurate dense correspondence, i.e., (1) a texture swapping constraint to ensure the same structure for images with the same structure code but different texture codes; (2) a texture swapping constraint to ensure similar texture for images with the same texture code, but different structure codes. We also introduce a warping loss to further regularize the correspondence maps. 
In addition, we show that CoordGAN can be flexibly equipped with an encoder that produces dense correspondence maps for real images.
We summarize our contributions as follows: 
\begin{itemize}
    \vspace{-3mm}
    \item We introduce a novel coordinate space from which dense correspondence across images of one category can be explicitly extracted. A warping function is introduced to learn this coordinate space.
    \vspace{-3mm}
    \item We propose CoordGAN, a disentangled GAN that generates dense correspondence maps and high-quality images, via a set of effective objectives.
    \vspace{-3mm}
    \item CoordGAN can be flexibly equipped with an encoder to produce the correspondence maps for real images. In other words, we also introduce a network (i.e., the encoder) that learns explicit structure representation.
    \vspace{-3mm}
    \item Experiments show that CoordGAN generates accurate dense correspondence maps and high-quality structure/texture editable images, for various categories.
\end{itemize}
\section{Related Work}

\textbf{Disentangled GANs.}
Recent studies~\cite{shen21closed,yang21semantic,harkonen2020ganspace} show that rich semantically meaningful directions (e.g., pose, color, lighting, etc.) automatically emerge in GANs. To factorize these meaningful latent directions, a line of disentangled GANs~\cite{nguyen-phuoc19hologan,chan21pi-gan,chen16infogan,singh19finegan,ojha21generating} are proposed to synthesize images via multiple latent factors, where each factor controls a certain attribute, e.g., object shape or texture.
Unlike~\cite{singh19finegan,ojha21generating, wang16generative} where human annotations (e.g., bounding boxes, surface normals, etc) are required, most related to ours are self-supervised disentanglement approaches~\cite{nguyen-phuoc19hologan, alharbiW20disentangled,kwon21diagonal}.
Among them, Alharbi et al.~\cite{alharbiW20disentangled} show that injecting hierarchical noise in the first layer of GANs leads to fine-grained spatial content disentanglement. Kwon et al.~\cite{kwon21diagonal} further inject noise into multiple layers with diagonal spatial attention modules. However, the learned content code only captures coarse structure such as viewpoints, i.e., keeping the same content code and only modifying the texture code would change the subject's shape. In contrast, our method models finer structure that allows for generating images of the same identity with various textures.

\textbf{Style Transfer.} Style transfer~\cite{choi18stargan, ulyanov16texture, Johnson16perceptual, gatys16image, park2020swapping} synthesizes a novel image by combining the content of one image with the texture of another one. Most related to ours is to swap texture between semantically-related regions of two images.
E.g., Park et al.~\cite{park2020swapping} learns a disentangled auto-encoder such that texture of corresponding regions can be swapped. In contrast, our work studies disentanglement of unconditional GANs and extracts dense correspondence between images explicitly.

\textbf{Dense Correspondence.} Identifying dense correspondence has been a challenging problem due to large shape and appearance variances. Most existing approaches are based on discriminative networks, i.e., either supervised image classification~\cite{choy2016universal,rocco2017convolutional,han2017scnet,jeon2018parn,semantic20liu,fcss17Seungryong}, or unsupervised image-level contrastive learning~\cite{xiao2021learning,xu2021rethinking}. Our work differs in that we investigate how to extract dense correspondence from GANs. Recently, several works~\cite{zhang21datasetgan,xu21linear} show that semantics can be extracted from GANs via a linear classifier in a few-shot setting. However, these methods still require manual annotations for training the classifier. Inspired by these works, we move one step further to extract dense correspondence without using any annotated labels.

\textbf{Concurrent Work.} Peebles et al.~\cite{peebles21gan} achieves visual alignment through equipping a pre-trained StyleGAN2~\cite{karras20analyzing} with additional Spatial Transformer Network (STN) ~\cite{jaderberg2015spatial}. 
However, dense correspondence is only identified for part of the object. 
Differently, through disentanglement of structure and texture, the proposed CoordGAN automatically generates correspondence maps of full images and neither pretrained StyleGAN nor additional STN is required.
\section{Dense Correspondence from CoordGAN}

\begin{figure}[t]
    \centering
    \includegraphics[width=0.7\linewidth]{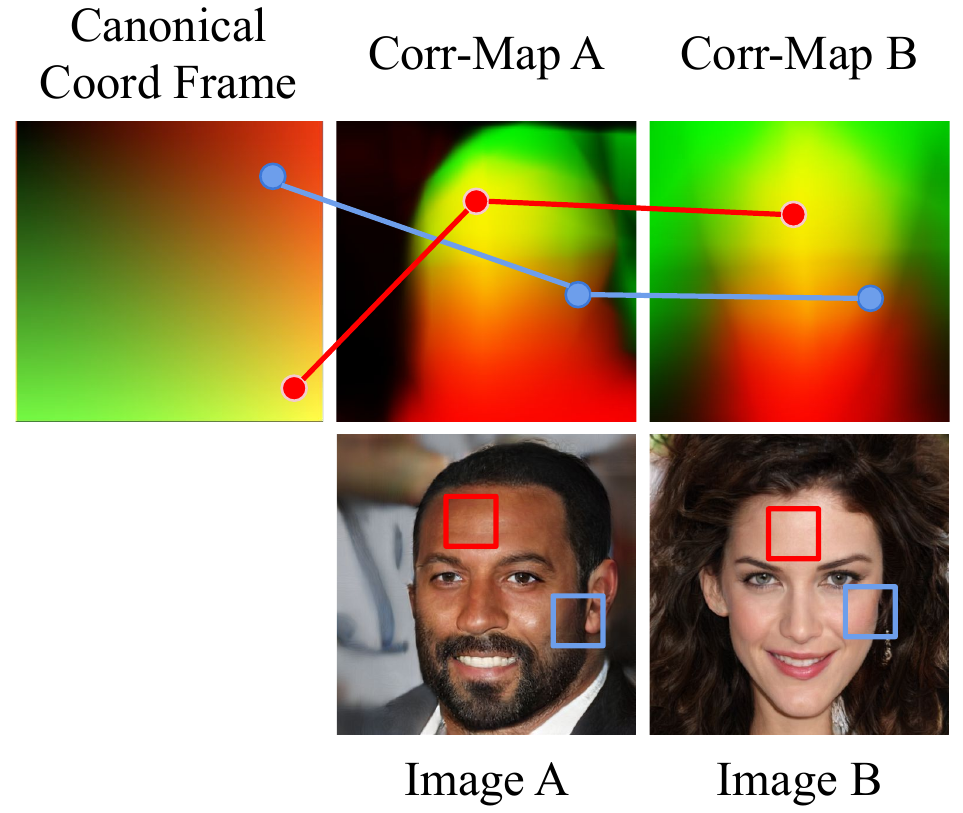}
    \vspace{-0.1in}
    \caption{Correspondence in coordinate space. The correspondence maps (Corr-Map) establish dense correspondence between all synthesized images and the canonical coordinate frame.}
    \label{fig:coord}
    \vspace{-0.15in}
\end{figure}

We design a structure-texture disentangled GAN such that dense correspondence can be extracted explicitly from the structural component, where the key component is to tie image structure to a coordinate space that is shared by all images.
Specifically, the structure of each generated image is represented as a warped coordinate frame, transformed from a shared canonical 2D coordinate frame. This reduces finding correspondence between image pixels to locating the coordinates in corresponding warped coordinate frames, which are transformed from the same coordinates in the canonical frame. We call our model Coordinate GAN (CoordGAN). 

\textbf{Coordinate Map Representation.} We define $C$ as a 2D coordinate map of width $W^c$ and height $H^c$. When $C(i,j) = (i,j)$, this denotes the {\it canonical coordinate map} (see Figure~\ref{fig:coord}). Pixel locations and coordinates are normalized to the range $[-1, 1]$. For example, $C(1, 1) = (1, 1)$ indicates the bottom right pixel of the coordinate map is of coordinate $(1, 1)$. It is then possible to define a {\it warping function\/} $\mathcal{W}: (C, w) \rightarrow C^{w}$, parameterized by a code $w \in \mathbb{R}^N$, that maps $C$ into a warped coordinate map $C^{w}$. 
Since the code $w$ relates the pixel coordinates of the image to the canonical coordinate map, it can be seen as the representation of image structure.
In particular, $C^w(i,j) = (k,l)$ implies that the pixel $i,j$ of the image is in correspondence with the canonical coordinate $k,l$. Given the two images with codes $w_1$ and $w_2$, it is also possible to establish correspondence between them by seeking pixels of similar coordinates. Given pixel $(i,j)$ of the image associated with coordinate $C^{w_1}(i,j)$, the corresponding pixel in another image of coordinate map $C^{w_2}$ is,
\begin{small}
\begin{equation}
    T_{1,2}(i,j) = \arg\min_{p,q} ||C^{w_1}(i,j) - C^{w_2}(p,q)||^2,
    \label{eq:transfer}
\end{equation}
\end{small}
where $T_{1,2}$ defines the forward transformation from warped coordinates $C^{w_1}$ to $C^{w_2}$. In this way, a generative model for images that includes a warping function automatically establishes dense correspondence between all synthesized images, as shown in Figure~\ref{fig:coord}. This can be useful for transferring properties between the images, such as semantic labels, landmark locations, image pixels, etc. 

\begin{figure*}[t]
    \centering
    \includegraphics[width=0.9\linewidth]{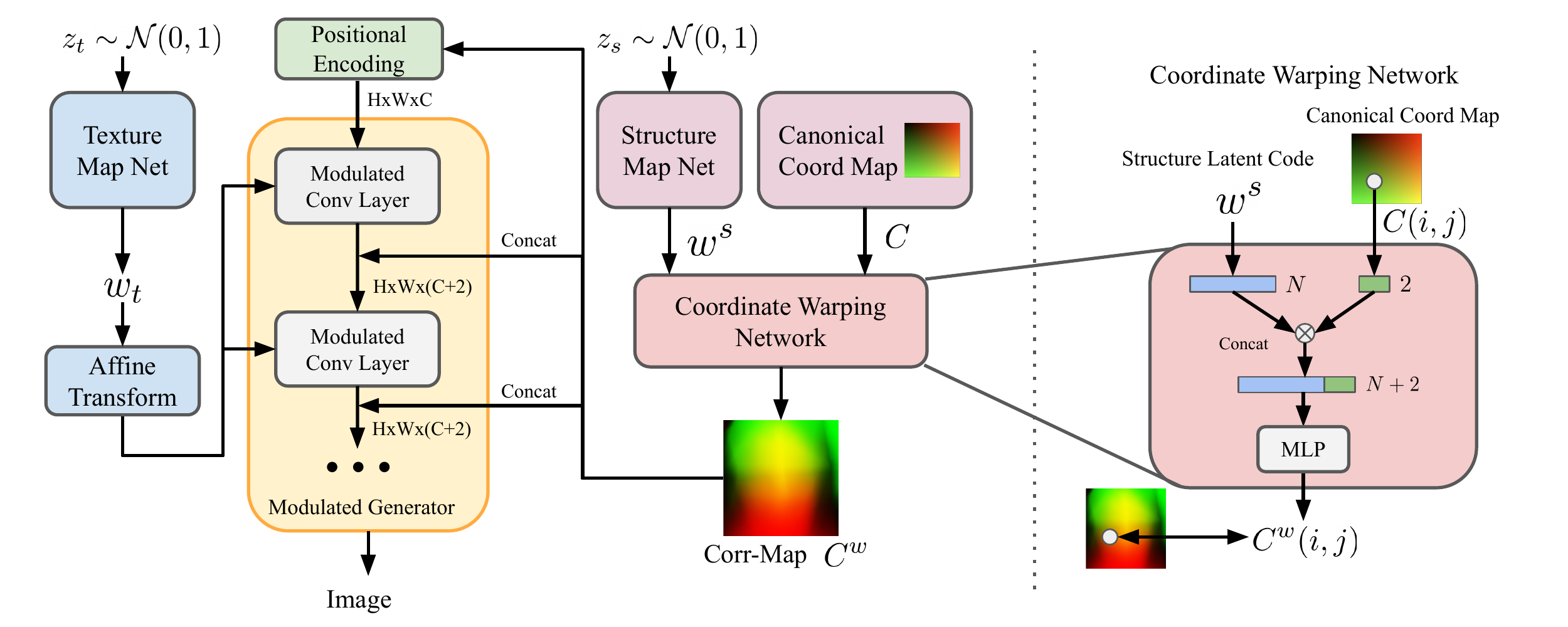}
    \vspace{-0.1in}
    \caption{Overview of CoordGAN. CoordGAN mainly consists of a texture mapping network, a structure mapping network, a coordinate warping network, and a modulated generator. The coordinate warping network (on the right) takes the structure latent code and a canonical coordinate map and outputs a correspondence map, which is then fed into multiple layers of the modulated generator to synthesize images.}
    \label{fig:arch}
    \vspace{-0.15in}
\end{figure*}

\subsection{Overview} \label{method:arch-overview}

An overview of the proposed CoordGAN is presented in Figure~\ref{fig:arch}. The CoordGAN is a generative model based on the structural coordinate map representation. The inputs to our model include two latent code vectors with dimension $N$: a structure code $z_{s}\in \mathcal{R}^N$ for modeling layouts and object structure, and a texture code $z_{t}\in \mathcal{R}^N$ for modeling texture, lighting, etc. The CoordGAN generator $G(z_{s}, z_{t}; \theta_{_G})$ is a mapping from these codes to the image space, with parameters $\theta_{_G}$. This is implemented by a combination of structure and texture mappings. A structure mapping network $w_s = {\cal S}(z_{s}; \theta_{\cal S})$ of parameters $\theta_{\cal S}$ maps the structure noise variable $z_s$ into a structure code $w_s$, which is then used by a warping function ${\cal W}(C,w_s)$ to produce a warped coordinate map $C^{w_s}$ for the image. A texture mapping network $w_t = {\cal T}(z_{t}, \theta_{\cal T})$ of parameters $\theta_{\cal T}$ maps the texture noise variable $z_{t}$ into a texture code $w_s$. The modulated generator then produces an image with the mapping $\cal A$ parameterized by $\theta_{\cal A}$,
\begin{small}
\begin{equation}
  G(z_{s}, z_{t}; \theta_{_G}) =  {\cal A}(C^{w_s}, w_t; \theta_{\cal A}),
\end{equation}
\end{small}
where $\theta_{_G}$ includes $\theta_{\cal S}, \theta_{\cal T},$ and $\theta_{\cal A}$.
The details of the various modules are discussed in the following sections.

\subsection{Coordinate Warping Network }\label{method:arch-warp}

One major component in CoordGAN is the warping function. We propose a Coordinate Warping Network, which learns a transformation between the canonical and a warped coordinate frame, conditioned on a latent structure code $w_s$. While there exist several differentiable transformation functions, such as Thin Plate Splines (TPS), Spatial Transformation Network (STN) \cite{jaderberg2015spatial}, and affinity matrix \cite{wang2018non,li2019joint}, in CoordGAN this transformation is implemented with a MLP as
\begin{small}
\begin{equation}
    C^{w_s}(i,j) = {\cal W}(C(i,j),w_s) = {\cal P}([C(i,j), w_s], \theta_{\cal P})~~~\forall i,j
\end{equation}
\end{small}
where ${\cal P}$ is a three layer MLP of parameters $\theta_{\cal P}$ and $[C(i,j), w_s]\in \mathcal{R}^{N+2}$ is the concatenation of coordinate $i,j$ from the canonical frame with the structure latent code $w_s$. In the supplementary materials, 
we show that the MLP is a learnable, conditional geometric transformation between the canonical coordinate frame and a warped coordinate frame.

The advantages of learning the transformation via a MLP are two folds.
First, since an MLP is a continues function containing only linear projection layers and ReLUs, it preserves the order of the coordinates in the canonical coordinate frame, i.e., it ensures that the warping is diffeomorphic.
Second, compared to TPS and STN, our design of ${\cal W}$ is generic and allows for more flexible deformation.

\subsection{Incorporating Warping in CoordGAN}\label{method:arch-gan}

We introduce the rest of CoordGAN components. While our generator design is inspired by StyleGAN~\cite{karras2019style} (see Figure. \ref{fig:arch}), we discuss the major differences in the following.


\textbf{Positional Encoding.} Rather than inputting dense correspondence map directly to the generator, we map it via a positional encoding layer~\cite{anokhin21image}. I.e., a Fourier embedding is obtained by the application of a $1 \times 1$ convolution followed by a sine function. The Fourier embedding serves as the first layer of the generator.


\textbf{Mapping Networks $\cal S(\cdot)$ and $\cal T(\cdot)$.} We use the same architecture as StyleGAN for the mapping network. Different from StyleGAN, we apply two independent mapping networks responsible for structure and texture, respectively.


\textbf{Modulated Generator ${\cal A}(\cdot)$.} 
We replace the learnable constant input of StyleGAN with the correspondence map. Since the latter has high resolution (i.e., $128\times 128$), instead of gradually increasing spatial resolution, the spatial resolution is kept the same as the input Fourier embedding at all layers as shown in Figure~\ref{fig:arch}. We inject the latent texture code $w_t$ into different layers of the modulated generator, via weight modulation~\cite{karras2020analyzing}, to render appearance details at different levels. To balance the structure and texture inputs at an architectural level, the dense correspondence map is also concatenated with the features produced by multiple intermediate layers of the modulated generator. We found that, without this multi-layer modulation of dense correspondence map, the coordinate warping network can only learn coarse and inaccurate structure information (e.g., viewpoints of faces), as shown in Table~\ref{table:ablation}.

\subsection{Learning Objectives}\label{method:loss}
To learn accurate correspondence maps and encourage the disentanglement of the latent space, such that $z_s$ and $z_t$ encode the image structure and texture separately, CoordGAN is trained with the following objectives.

\textbf{Texture Swapping Constraint.} To ensure the CoordGAN generates the same identity and image layout when the structure is fixed and only the texture code is modified, a texture swapping constraint is applied. Given a pair of synthesized images with a shared structure code $z_s$ and different texture codes $z_{t_1}, z_{t_2}$, the texture swapping loss $\mathcal{L}_{t}$ is defined as the LPIPS~\cite{zhang18lpips} loss between the two synthesized images:
\begin{small}
\begin{equation}
    \mathcal{L}_{t} = \mathcal{L}_{_{LPIPS}}(G(z_s, z_{t_1}; \theta_{_G}), G(z_s, z_{t_2}; \theta_{_G})).
\end{equation}
\end{small}

\textbf{Structure Swapping Constraint.} To encourage images that share the same texture code to have similarly textures, a structure swapping constraint is introduced. This consists of encouraging two images with the same texture code $z_t$ but different structure codes $z_{s_1}$ and $z_{s_2}$ to have similar textures. Following~\cite{park2020swapping}, this is done with a  non-saturating GAN loss based on a patch discriminator $D_{patch}$:
\begin{small}
\begin{equation}
    \mathcal{L}_{s} = \mathop{\mathbb{E}} \Big[ - \log \Big( D_{patch} \big(G(z_{s_1}, z_{t}; \theta_{_G}), G(z_{s_2}, z_{t}; \theta_{_G}) \big) \Big) \Big].
\end{equation}
\end{small}

\textbf{Warping Loss.} 
A warping loss is defined to explicitly regularize the correspondence map. Given a pair of synthesized images $x_{_1} = G(z_{s_1}, z_{t_1}; \theta_{_G})$ and $x_{_2} =  G(z_{s_2}, z_{t_2}; \theta_{_G})$, $x_{_1}$ is warped to the coordinate frame of $x_{_2}$ by transferring pixel colors according to Equation~(\ref{eq:transfer}). In practice, similar to \cite{wang19learning,wang2018non,li2019joint}, we relax Equation~(\ref{eq:transfer}) with affinity matrix to make the warping differentiable. This produces a warped image $x_{_{2,1}}^w$. A warping loss based on the LPIPS loss~\cite{zhang18lpips},
\begin{small}
\begin{equation}
    \mathcal{L}_{warp} = \mathcal{L}_{_{LPIPS}}(x_{_{2,1}}^w, x_{_{2}}),
\end{equation}
\end{small}
is used to minimize the distance between $x_{_{2,1}}^w$ and $x_{_{2}}$ .

\textbf{Chamfer Loss.} Suppose a canonical coordinate map $C$ is transformed to a warped coordinate map $C^w$, a Chamfer loss is implemented to avoid the collapse of the transformation,
\begin{small}
\begin{equation}
    \begin{aligned}
        \mathcal{L}_{cham} = & \frac{1}{|C|}\sum_{(i,j) \in C} \min_{(p,q)} ||C(i,j) - C^{w}(p,q)||_{2} \\
        & + \frac{1}{|C^{w}|}\sum_{(p,q) \in C^{w}} \min_{(i,j)} || C^{w}(p,q) - C(i,j) ||_{2}.
    \end{aligned}
\end{equation}
\end{small}

\textbf{Overall Learning Objective.} To generate realistic images, a standard GAN objective function $\mathcal{L}_{GAN}$ is applied to the synthesized images. Combining all the aforementioned loss objectives, the overall training objective is defined as 
\begin{small}
\begin{equation}
    \begin{aligned}
        \mathcal{L}_{G} =& \lambda_{t} * \mathcal{L}_t + \lambda_{s} * \mathcal{L}_s + \lambda_{warp} * \mathcal{L}_{warp} \\
                         & + \lambda_{cham} * \mathcal{L}_{cham} + \lambda_{GAN} * \mathcal{L}_{GAN},
    \end{aligned}
\end{equation}
\end{small}
where $\lambda_{t}, \lambda_{s}, \lambda_{warp}, \lambda_{cham}, \lambda_{GAN}$ are coefficients used to balance the different losses.

\subsection{Inverting CoordGAN via an Encoder}\label{method:encoder}
The CoordGAN can be equipped with an encoder to enable the extraction of dense correspondence from real images. Specifically, an encoder $E(\cdot;  \theta_{_E})$ parameterized by $\theta_{_E}$ is introduced to map an image $x$ to a pair of structure  $w_{s,E}$ and texture $w_{t,E}$ latent codes. These latent codes are then input to the CoordGAN to synthesize a replica of the image.  As observed in \cite{richardson21pix2style2pix}, embedding real images directly into $W+$ space rather than $W$ space leads to better reconstruction. So for the texture branch, we design the encoder to output texture latent codes $w_{t,E}^{+}$ in $W+$ space as opposed to $w_{t,E}$ in $W$ space. During training, we fix the generator while optimizing the encoder via latent consistency, reconstruction and texture swapping losses, which are described as follows.

\textbf{Latent Consistency Loss.} We introduce a latent consistency loss by feeding synthesized images back to the encoder and matching the distribution of encoder outputs to that originally produced by the mapping network.  
Suppose an image is synthesized with latent codes $w_t$, $w_s$, and correspondence map $C^w$. Inputting this image back into the encoder produces a pair of latent codes $w_{t,E}^+$ and $w_{s,E}$, and the correspondence map $C^w_{E}$. The latent consistency loss $\mathcal{L}_{con}$ is defined as
\begin{small}
\begin{equation}
    \mathcal{L}_{con} = \mathcal{L}_2(w_s, w_{s,{_E}}) + \mathcal{L}_2(C^w, C^w_{_E}),
\end{equation}
\end{small}
where $\mathcal{L}_2(\cdot, \cdot)$ denotes the $\mathcal{L}_2$ loss.

\textbf{Reconstruction Loss.} This is a reconstruction loss for input real images, with L1 ($\mathcal{L}_1$) and LPIPS~\cite{zhang18lpips} ($\mathcal{L}_{LPIPS}$) components, defined as

\begin{small}
\begin{equation}
    \mathcal{L}_{rec} = \mathcal{L}_1(x, G(E(x))) + \mathcal{L}_{_{LPIPS}}(x, G(E(x))),
\end{equation}
\end{small}

\textbf{Overall Learning Objective.} The overall learning objective used for encoder training an encoder is 
\begin{small}
\begin{equation}
    \begin{aligned}
        \mathcal{L}_{E} = \lambda_{con} * \mathcal{L}_{con} + \lambda_{rec} * \mathcal{L}_{rec} + \lambda_{t} * \mathcal{L}_{t},
    \end{aligned}
\end{equation}
\end{small}
where $\lambda_{con}, \lambda_{rec}, \lambda_{t}$ are hyperparameters that balance the different losses.

We note that the encoder facilitates explicit structure representation learning for real images. It is significantly more efficient than optimization-based GAN-inversion methods, as no iterative inference is required.

\section{Experiments}

In this section, we show quantitative and qualitative results of models trained on the CelebAMask-HQ~\cite{CelebAMask-HQ}, Stanford Cars~\cite{li13stanfordcar}, and AFHQ-Cat~\cite{choi18stargan} datasets. We train separate models on each dataset, using a  resolution of $512 \times 512$ for the CelebAMask-HQ model and $128 \times 128$ for the other two. For CelebAMask-HQ, we first train CoordGAN with an output size of $128 \times 128$ and then append two upsampling layers to generate high-resolution images. Detailed network design and training hyper-parameters are described in the supplementary.

\begin{figure*}[t]
    \centering
    \includegraphics[width=\linewidth]{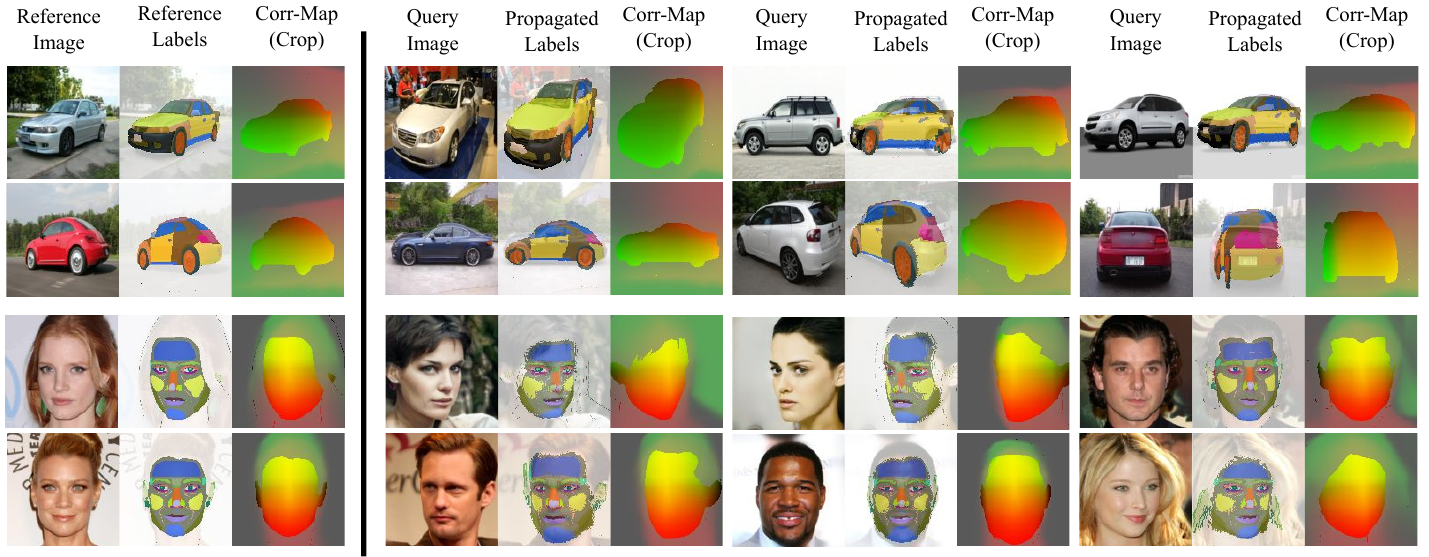}
    \vspace{-0.30in}
    \caption{Qualitative results for semantic label propagation. In each row, given one reference image along with its semantic labels as shown on the left, the proposed approach predicts its correspondence map and propagates its segmentation mask to other query images on the right. For better visualization, we use the ground-truth masks to highlight the foreground areas of all the predicted correspondence maps, denoted with Corr-Map (Crop). Note that no ground-truth masks are used for actual label propagation.}
    \label{fig:correspondence}
    \vspace{-0.15in}
\end{figure*}

\subsection{Evaluation on Dense Correspondence}\label{exp:correspondence}
We quantitatively demonstrate the quality of the extracted dense correspondence on the task of semantic label propagation. Given one reference image with semantic labels, its correspondence map is first inferred with the trained encoder. This establishes a mapping between the semantic labels and the correspondence map for that image. Another correspondence map is then inferred for a query image and the labels of the reference image are obtained with Equation~(\ref{eq:transfer}). To align with the training stage, we relax Equation~(\ref{eq:transfer}) with affinity matrix in practice.

\textbf{Datasets and Metrics.} We evaluate different methods on the CelebAMask-HQ~\cite{CelebAMask-HQ} and DatasetGAN~\cite{zhang21datasetgan} datasets. 
We merge CelebAMask-HQ dataset labels and select 6 classes (eyes, nose, ear, mouth, face and eyebrow) for our evaluation. The DatasetGAN dataset contains detailed manually annotated labels for faces (34 classes) and cars (20 classes). 
For the DatasetGAN faces, we excluded neck and hair since they are not consistently visible for all images in the dataset.
For all datasets, we randomly select 5 images as reference and another set as query images. Each reference image's semantic label is propagated to all query images and the mean intersection-over-union (IOU) with the ground-truth segmentation maps is computed for evaluation. We report the averaged score of these 5 runs.

\textbf{Baselines.} For all baseline models, we extract features from hidden layers and use nearest neighbor search to determine feature correspondences and propagate labels. We detail the features selected for label propagation below.
We employ two sets of baselines. The first set comprises of transfer learning based methods with either supervised ImageNet pre-training, e.g., ResNet50~\cite{he16resnet} or self-supervised contrastive learning based pre-training, e.g., MoCo~\cite{moco20kaiming} pre-trained on ImageNet~\cite{deng09imagenet} and VFS~\cite{xu2021rethinking} pre-trained on Kinetics video dataset~\cite{kay17kinetics}. For all these methods, ResNet50~\cite{he16resnet} is employed as the backbone and the pre-trained models are directly tested on our task without fine-tuning. We follow~\cite{xiao2021learning, xu2021rethinking} and use the Res-block 4 features for label propagation as it is shown that Res-block 4 gives the best pixel-level correspondences. Another set of baselines is based on auto-encoders, such as Swapping Auto-encoder~\cite{park2020swapping} and Pix2Style2Pix~\cite{richardson21pix2style2pix}. Both methods are trained on the same datasets as ours. 
For Swapping Auto-encoder, the structure branch features are used for label propagation. For Pix2Style2Pix encoder, the Res-block 4 features are used for label propagation. 
All methods are evaluated with input image resolution of 128, except for Pix2Style2Pix where the input image size is set to 256 following the original paper.

    \begin{table}[t]
    \fontsize{8}{10pt}\selectfont
    \setlength{\tabcolsep}{3pt}
    \centering
    \begin{tabular}{l|cccc} 

    \multirow{3}{*}{} & CelebA-HQ & DGAN-face & DGAN-car \\
    \shline
    Resnet50~\cite{he16resnet}  & 39.48 & 11.05 & 11.07\\
    \shline
    Moco~\cite{moco20kaiming}  & 36.19 & 10.00 & 9.53\\
    VFS~\cite{xu2021rethinking}  & 38.10 & 8.55 & 6.88\\
    Swap AE~\cite{park2020swapping} & 24.73 & 5.48 & 5.37 \\
    Pix2Style2Pix~\cite{richardson21pix2style2pix} & 48.50 & 20.36 & 10.77 \\
    CoordGAN & \bf 52.25 & \bf 23.78 & \bf 13.23 \\

    \end{tabular}
        \vspace{-0.10in}
    \caption{IOU comparison for label propagation. Our method shows the best semantic label propagation results among all baseline methods.}
        \vspace{-0.30in}
    \label{table:correspondence}
    \end{table}

\begin{figure*}[t]
    \centering
    \includegraphics[width=\linewidth]{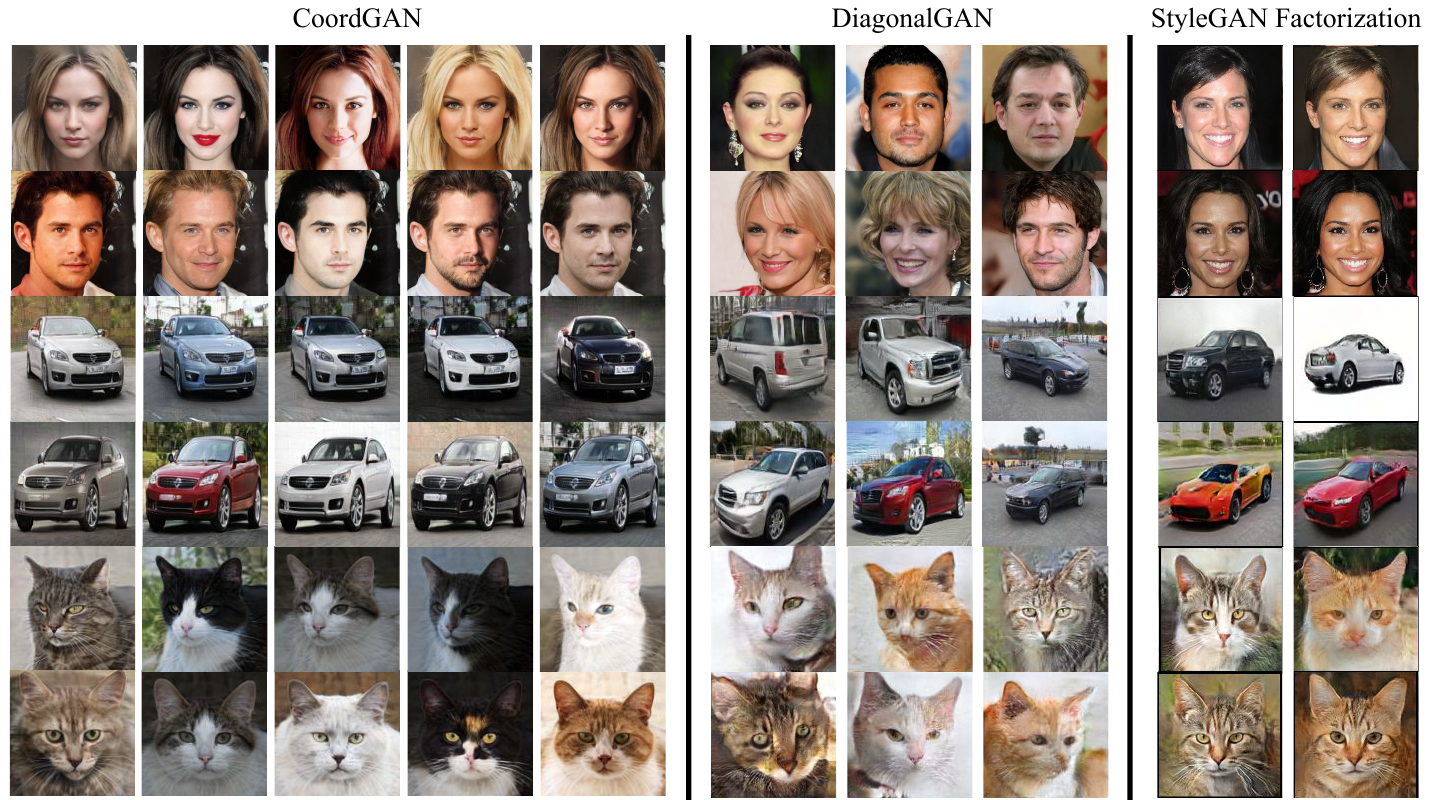}
    \vspace{-0.30in}
    \caption{Qualitative comparison for texture swapping. 
    From top to bottom: models trained on CelebAMask-HQ, Stanford Cars, and AFHQ-cat datasets. 
    For CoordGAN and DiagonalGAN, images shown in each row are generated with the same structure code and diverse texture codes. For GAN Factorization, images in each row are generated with random perturbations along the identified eigen-vector directions. It is apparent that CoordGAN preserves structure better when only texture codes are modified.}
    \label{fig:texture_swap}
    \vspace{-0.2in}
\end{figure*}

\textbf{Quantitative Results}. As reported in Table~\ref{table:correspondence}, the proposed CoordGAN outperforms all baselines across all three datasets on the task of semantic segmentation label propagation. The most related approach is Pix2Style2Pix, which also learns an encoder for a pre-trained StyleGAN2 model. While Pix2Style2Pix encoder features contain both structure and texture information, CoordGAN correspondence maps, with only structure information, still achieve better label propagation performance. These results suggest that CoordGAN learns much accurate correspondence than the other methods.

\textbf{Qualitative Results.} We visualize both the coordinate maps and the propagated segmentation labels in Figure~\ref{fig:correspondence}. On the left, several reference images from the DatasetGAN dataset are shown along with their semantic labels. On the right, we show the propagation results for different query test images. The predicted correspondence maps for both the reference and query images are color-coded and masked with the foreground ground-truth semantic labels for better visualization. Note that this is only for visualization, no ground-truth masks are used for the actual label propagation. Note that our method produces precise label propagation results for both frontal and profile query faces. For cars, this is even more challenging, considering the large differences in viewpoints and scales. For example, in extreme cases where the reference car is viewed from the front and the query car from the back, no correspondence exist. Surprisingly, even in cases where the reference car is observed from the side and the query car from the rear, CoordGAN still matches the labels reasonably well. We conjecture this is because it learns a reasonable prior for the category, by observing many instances and densely associating them during training.

\subsection{Identity-preserving Texture Swapping}\label{exp:texture_swap}

We analyze disentanglement of structure and texture of CoordGAN by generating images with the same structure code but different texture codes (i.e., texture swapping) and evaluating the structural consistency of the outputs. We focus on the generator and do not use an encoder in these experiments.

    \begin{table}[t]
    \fontsize{7}{10pt}\selectfont
    \setlength{\tabcolsep}{2pt}
    \centering
    \begin{tabular}{l|ccc|cc|cc} 
     &\multicolumn{3}{c|}{CelebA-HQ} & \multicolumn{2}{c|}{Stanford Cars} & \multicolumn{2}{c}{AFHQ-cat}  \\
     & LPIPS $\downarrow$ & Arcface $\downarrow$ & FID $\downarrow$ & LPIPS $\downarrow$ & FID $\downarrow$ & LPIPS $\downarrow$ & FID $\downarrow$\\
    \shline
    StyleGAN2~\cite{karras20analyzing}  & - & - & 8.21 & - & 16.20 & - & 21.02\\
    DiagonalGAN~\cite{kwon21diagonal}  & 0.58 & 0.79 & 11.16 & 0.61 & 18.09 & 0.55 & 17.63\\
    CoordGAN & \bf 0.22 & \bf 0.38 & 16.16 & \bf 0.21 & 24.27 & \bf 0.27 & 23.62\\

    \end{tabular}
        \vspace{-0.10in}
    \caption{Texture swapping comparison. The lowest LPIPS and Arcface feature distances of CoordGAN suggest  better structure preservation when the texture code is varied.}
        \vspace{-0.25in}
    \label{table:GAN}
    \end{table}

\textbf{Metrics.} To quantitatively examine different methods, we use the ArcFace~\cite{deng19arcface} face identity loss and the LPIPS~\cite{zhang18lpips} loss to evaluate disentanglement and structure preservation performance, and FID~\cite{heusel17fid} score for measuring the perceptual image quality of the generated images. 
ArcFace computes a feature-level cosine similarity loss between two faces. It can be used to measure whether the face identity is preserved since the smaller the loss is, the more likely both images capture the same identity. LPIPS~\cite{zhang18lpips} measures whether two images have similar image layouts.

\textbf{Baselines.} CoordGAN is compared against two baselines: DiagonalGAN~\cite{kwon21diagonal} and  GAN Factorization~\cite{shen21closed}. DiagonalGAN achieves state-of-the-art performance for StyleGAN-based structure and texture disentanglement. Similar to CoordGAN, it uses separate structure and texture codes as inputs. To generate texture-swapped images, we sample a structure code and different texture codes, and then compute the structural similarity  among the images synthesized using the aforementioned metrics. GAN Factorization exploits SVD to identify semantically meaningful latent directions across different GAN layers. The paper suggests that the final layers of the GAN are mainly responsible for controlling texture. Therefore, we generate texture-swapped images with GAN Factorization by adding perturbations along the computed eigen-vectors of the last two convolution layers of a pre-trained StyleGAN2.

\textbf{Results.} As shown in Table~\ref{table:GAN}, CoordGAN outperforms the baselines by a significant margin for all disentanglement metrics (ArcFace and LPIPS) on all object categories. This suggests that it successfully preserves the fine-grained image structure independent of the input texture. Note that ArcFace is only available for human faces. The FID score is computed over 10,000 generated images for all methods, for reference. Note that, as discussed in~\cite{kwon21diagonal,alharbiW20disentangled}, a slight decrease in the FID score is observed due to the strong disentanglement constraints enforced.

In Figure~\ref{fig:texture_swap}, each row shows diverse texture-swapped images generated by fixing the structure code and varying the texture code.  The DiagonalGAN changes the subject's identity completely. This becomes more clear when testing on cars, where the viewpoint is ambiguous and scale can vary. Results suggest that its disentangled content code only captures coarse structural information, such as the rough image layout and viewpoint. 
In contrast, CoordGAN successfully maintains both the coarse and fine-grained image structure and only varies appearances, on all datasets. 
For GAN factorization, while potentially possible to exhaustively search for the latent eigen-vectors that only modify image textures, it is not easy to finely control the appearance of the synthesized images. 

\begin{figure}[t]
    \centering
    \includegraphics[width=\linewidth]{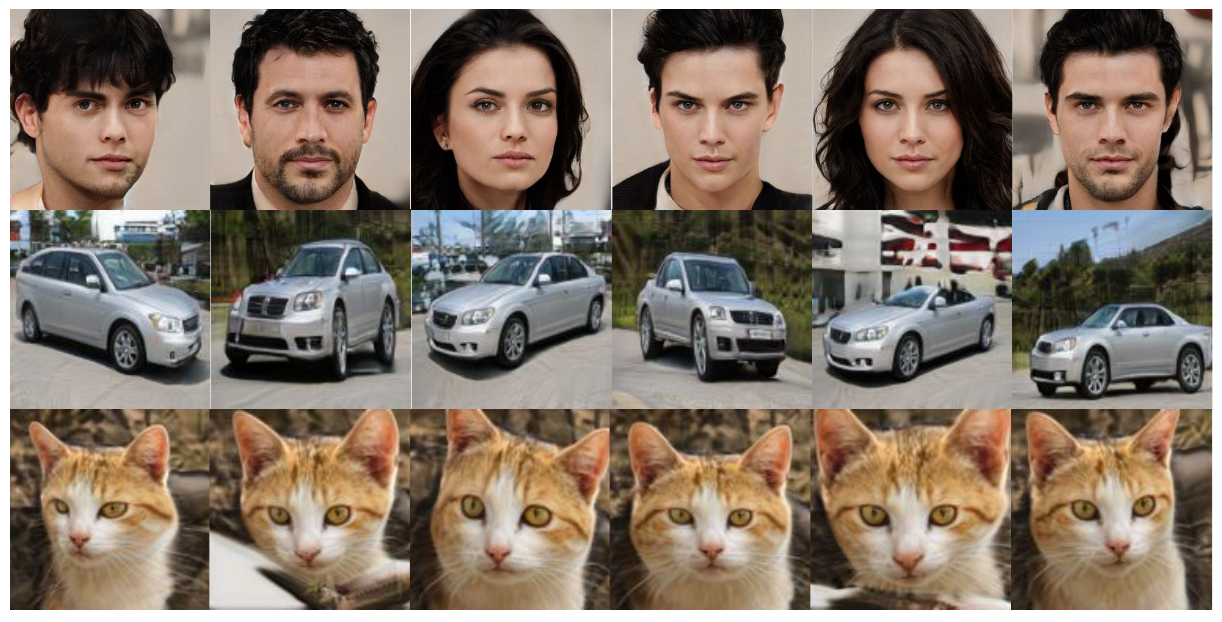}
    \vspace{-0.20in}
    \caption{Qualitative results for structure swapping. 
    Images shown in each row are generated with the same texture code and diverse structure codes.}
    \label{fig:structure_swap}
    \vspace{-0.25in}
\end{figure}
\subsection{Structure Swapping}\label{exp:structure_swap}
To further demonstrate CoordGAN successfully disentangles structure and texture, in this section, we synthesize images of the same texture code and various structure codes (i.e., structure swapping). As show in Figure~\ref{fig:structure_swap}, from top to bottom, we show synthesized images of models trained separately on CelebAMask-HQ, Stanford Cars, and AFHQ-cat datasets. It is clear that images in each row show similar textures (e.g., hair/face colors for humans, grey cars, orange cats) with diverse structural variations (e.g., viewpoint, scale, shape, layout, etc). The again confirms that CoordGAN learns a disentangled representation where the structure code and the texture code capture different attributes of a synthesized image. More visualizations are included in the supplementary materials.

\subsection{Ablation Studies}\label{exp:alation}
We ablate different architectures w.r.t the structure branch, i.e., feeding the correspondence map (1) only to the first layer of CoordGAN  (w/o struc-mod), or (2) to modulate multiple layers, as discussed in Section~\ref{method:arch-gan}.  Both models are trained to synthesize images of resolution $128 \times 128$ on the CelebAMask-HQ dataset. 
Table~\ref{table:ablation} shows that the proposed structure modulation design is crucial to achieve a good disentanglement of structure and texture.
This confirms that a non-trivial architecture design is needed to embed the structure information and highlights the importance of the proposed balanced structure and texture modulation. More studies on objectives are included in the supplementary materials.

    \begin{table}[t]
    \fontsize{7}{10pt}\selectfont
    \setlength{\tabcolsep}{3pt}
    \centering
    \begin{tabular}{l|cc|cc} 
     &\multicolumn{2}{c|}{Disentanglement} & \multicolumn{2}{c}{Correspondence} \\
     & LPIPS$\downarrow$ & Arcface $\downarrow$ & CelebA-HQ & DGAN-face \\
    \shline
    CoordGAN  & 0.10 & 0.32 & 52.25 & 23.78 \\
     \hspace{0.5cm} w/o struc-mod  & 0.32 & 0.73 & 48.59 & 20.01\\

    \end{tabular}
    \vspace{-0.1in}
    \caption{Ablation on structure modulation. We show that incorporating the structure modulation is essential to a good disentanglement and correspondence performance (measured by IOU).}
    \vspace{-0.2in}
    \label{table:ablation}
    \end{table}

\vspace{-0.05in}
\section{Discussion}

\vspace{-0.05in}
\paragraph{Conclusion. }
    In this work, we show that it is possible to train GANs so that dense correspondence can automatically emerge. We propose a novel disentangled GAN model, CoordGAN, that produces dense correspondence maps represented by a novel coordinate space. This is complemented by an encoder for GAN inversion, which enables the generation of dense correspondence for real images. Experimental results show that CoordGAN generates accurate dense correspondence maps for a variety of categories. This opens up a new door for learning dense correspondences from generative models in an unsupervised manner. We qualitatively and quantitatively demonstrate that CoordGAN successfully disentangles the structure and texture on multiple benchmark datasets. 

\vspace{-0.15in}
\paragraph{Limitations and Future Work.} The current proposed model is restricted to learn correspondence within the same category, since it requires the coordinate maps transformed from the same canonical space. While we can potentially infer the 3D viewpoints from the coordinate map (as visualized in Figure~\ref{fig:correspondence}), we have not explicitly modelled the 3D structure in our representation. A future extension of this work can be to learn a 3D UV coordinate map instead of a 3D map to represent the underlying structure. 



\vspace{1em}
{\footnotesize \textbf{Acknowledgements.}~This work was supported, in part, by grants from NSF IIS-1924937, NSF IIS-2041009, NSF CCF-2112665 (TILOS), and gifts from Qualcomm and Amazon.}

{\small
\bibliographystyle{ieee_fullname}
\bibliography{egbib}
}

\newpage
\appendix

\section{Overview}
In this appendix, we provide more details of the submission: We show quantitative results of structure swapping in Section \ref{sec:struc swap}; We provide ablation studies on the objectives, as described in Section \ref{sec:abl}; We conduct a user study on disentangling structure and texture by swapping the attributes of the generated images, as described in Section \ref{sec:dis}; We answer the question of how an MLP models explicit transformation between canonical and warped coordinate frames in Section \ref{sec:backward-mlp}; More implementation details are discussed in Section \ref{sec:imp}; We show at last in Section \ref{sec:metface} the application of the finetuning the CoordGAN on other domains.

\section{Quantitative Results on Structure Swapping}\label{sec:struc swap}
We present quantitative comparisons to the SOTA  structure swapping method DiagonalGAN.
We sample 5000 pairs of images, each pair with the same texture code and different structure codes.
Each pair is evaluated by both the
\textit{LPIPS} (to measure how the structure varies) and the \textit{ArcFace} (to evaluate whether identity changes) scores. 
CoordGAN has better performance: \textbf{0.75 over 0.65 for ArcFace}, and \textbf{0.55 over 0.50 for LPIPS}. 
The results indicate that, with the same texture code, different structure codes of CoordGAN produce images of larger structural variations, whereas DiagonalGAN tends to generate images with similar identities. This demonstrates the structure and texture are better disentangled with the proposed method.

\section{Ablation Studies} \label{sec:abl}

\textbf{Ablation for image synthesis.}
In this part, we study the effect of different loss functions for training the generator. As shown in Table~\ref{table:ablation-generator}, it is observed that the combination of the warp loss, structure swapping constraint and texture swapping constraint achieve the best performance. Without the texture swapping constraint, the texture code tends to take the majority of the variances while the warped coordinates are similar across different samples. Without the warp loss, the correspondence performance drops significantly. This suggests that regularizing the correspondence maps is crucial to extracting dense correspondence accurately.

\textbf{Ablation for encoder.}
In this part, we fix the parameters of the generator and study the effects of different loss functions for training the encoder. Table~\ref{table:ablation-encoder} compares the reconstruction performance with respect to different loss combinations. We find that both the latent consistency and the texture loss are essential to achieving the best reconstruction performance. While removing the texture swapping loss results in lower reconstruction errors, we find the correspondence performance slightly decreases. Without the latent consistency loss, both reconstruction and correspondence performance drop significantly. This indicates that encouraging the encoded structure to match the learned distribution plays an important role to model accurate correspondence for real image inputs.

\begin{figure*}[t]
    \centering
    \includegraphics[width=\linewidth]{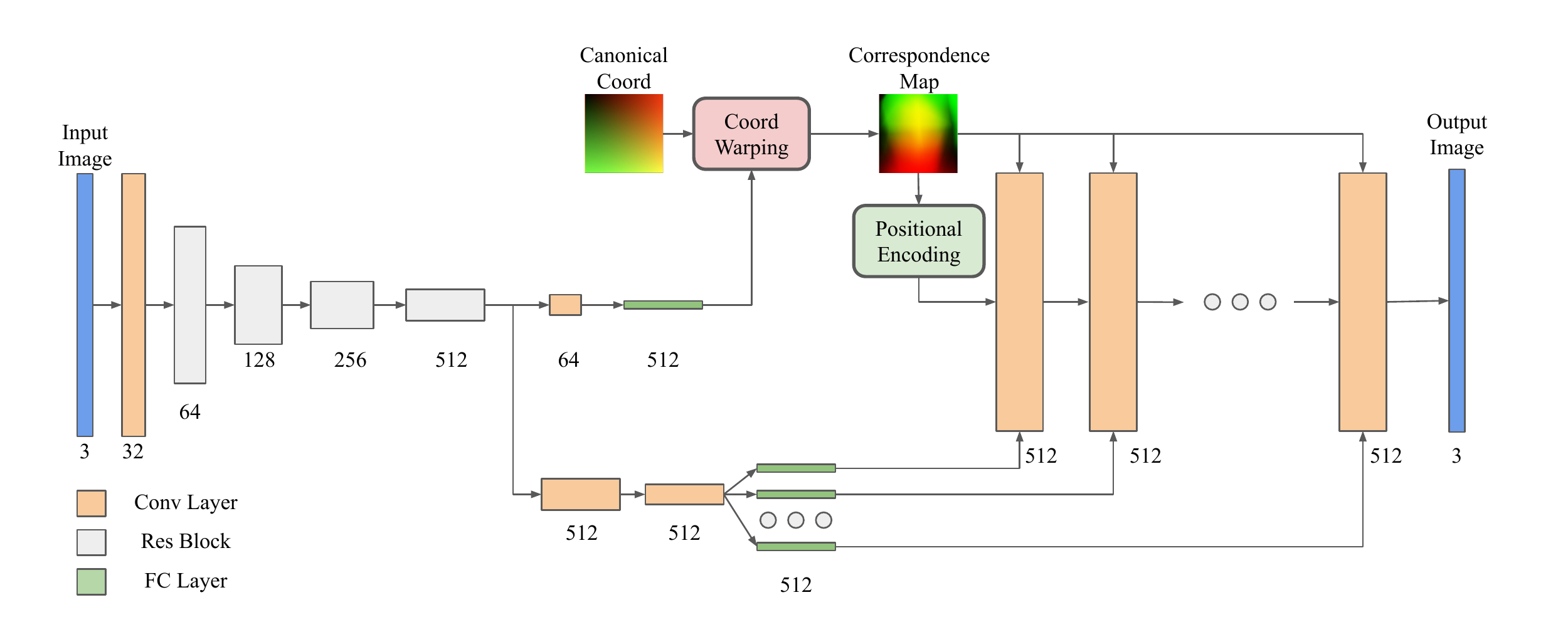}
    \caption{Auto-encoder architecture. The encoder takes an image as input and outputs a latent structure code and a latent texture code. Then the generator takes the predicted structure and texture latent codes and outputs images. }
    \label{fig:encoder}
\end{figure*}

    \begin{table}[t]
    \fontsize{8}{10pt}\selectfont
    \setlength{\tabcolsep}{3pt}
    \centering
    \begin{tabular}{l|cc|cc} 
     &\multicolumn{2}{c|}{Disentanglement} & \multicolumn{2}{c}{Label Propagation} \\
     & LPIPS$\downarrow$ & Arcface $\downarrow$ & CelebA-HQ & DGAN-face \\
    \shline
    CoordGAN  & 0.10 & 0.32 & 52.25 & 23.78 \\
    \hspace{0.1cm} w/o warp loss  & 0.11 & 0.22 & 24.84 & 10.52\\
    \hspace{0.1cm} w/o structure swap  & 0.18 & 0.51 & 46.52 & 18.53\\
    \hspace{0.1cm} w/o texture swap  & 0.64 & 0.90  & 45.96 & 17.66\\
    \end{tabular}
    \caption{Ablation on generator losses on CelebAMask-HQ. We show that incorporating the all losses is essential to good disentanglement and label propagation performance (measured by IOU).}
    \label{table:ablation-generator}
    \end{table}

    \begin{table}[t]
    \fontsize{7}{10pt}\selectfont
    \setlength{\tabcolsep}{3pt}
    \centering
    \begin{tabular}{l|ccc|cc} 
     &\multicolumn{3}{c|}{Reconstruction} & \multicolumn{2}{c}{Label Propagation} \\
     & LPIPS$\downarrow$ & Arcface $\downarrow$ & MSE$\downarrow$ & CelebA-HQ & DGAN-face \\
    \shline
    CoordGAN  & 0.25 & 0.49 & 0.03 & 52.25 & 23.78\\
    \hspace{0.1cm} w/o latent consistency  & 0.28 & 0.59 & 0.04 & 46.19 & 21.50\\
    \hspace{0.1cm} w/o texture swap loss  & 0.23 & 0.47 & 0.03 & 50.83 & 23.53\\
    \end{tabular}
    \caption{Ablation on encoder losses on CelebAMask-HQ. We show that incorporating all the losses is essential to faithfully reconstructing the input and encoding accurate correspondence.}
    \label{table:ablation-encoder}
    \end{table}

\section{User Study on Attribute Swapping}\label{sec:dis}
We conduct a user study to further evaluate the disentanglement of structure and texture for the proposed CoordGAN and DiagonalGAN. Given a pair of images generated with the same structure code but diverse texture codes, we ask users on AMT to rate the pairs of images based on their structural similarity with a score from 1 to 5. A higher score indicates that the pair of images are more similar in terms of structure. 
%
Likewise, we ask users to rate the texture similarities of images generated with the same texture code but diverse structure codes.
For each dataset, we randomly sample 200 image pairs and each pair of images is rated independently by three individuals.

\begin{figure*}[t]
    \centering
    \includegraphics[width=\linewidth]{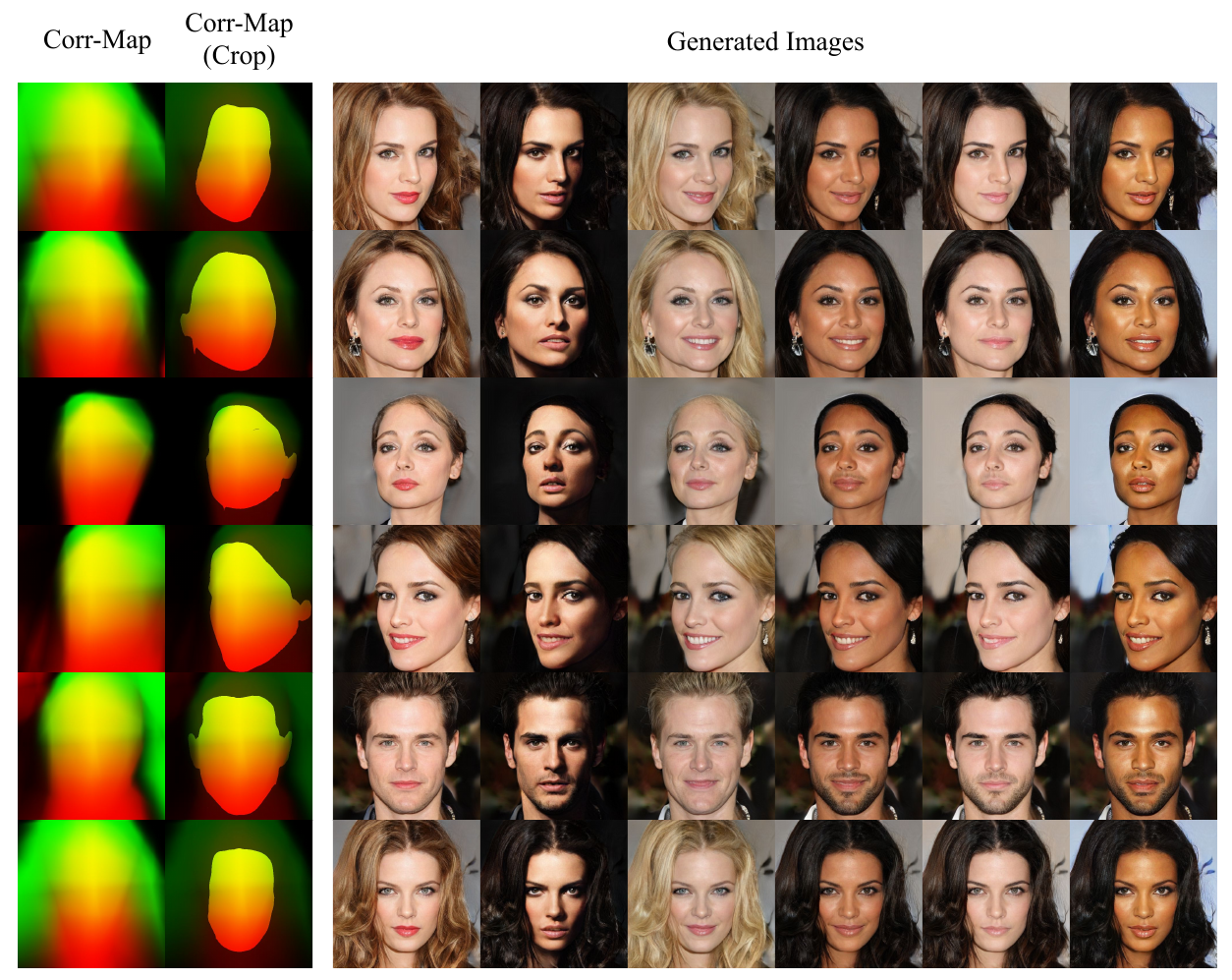}
    \vspace{-0.30in}
    \caption{Images synthesised by the proposed CoordGAN model: each row displays images with the same structure but different textures; in each column, structure varies while texture is fixed. The correspondence maps (Corr-Map) controlling the structure of the synthesized images are shown in the first column of each row. For better visualization, we use off-the-shelf segmentation models to highlight the foreground areas of all the predicted correspondence maps, as shown with Corr-Map (Crop).}
    \label{fig:face1}
    \vspace{-0.15in}
\end{figure*}

\begin{figure*}[t]
    \centering
    \includegraphics[width=\linewidth]{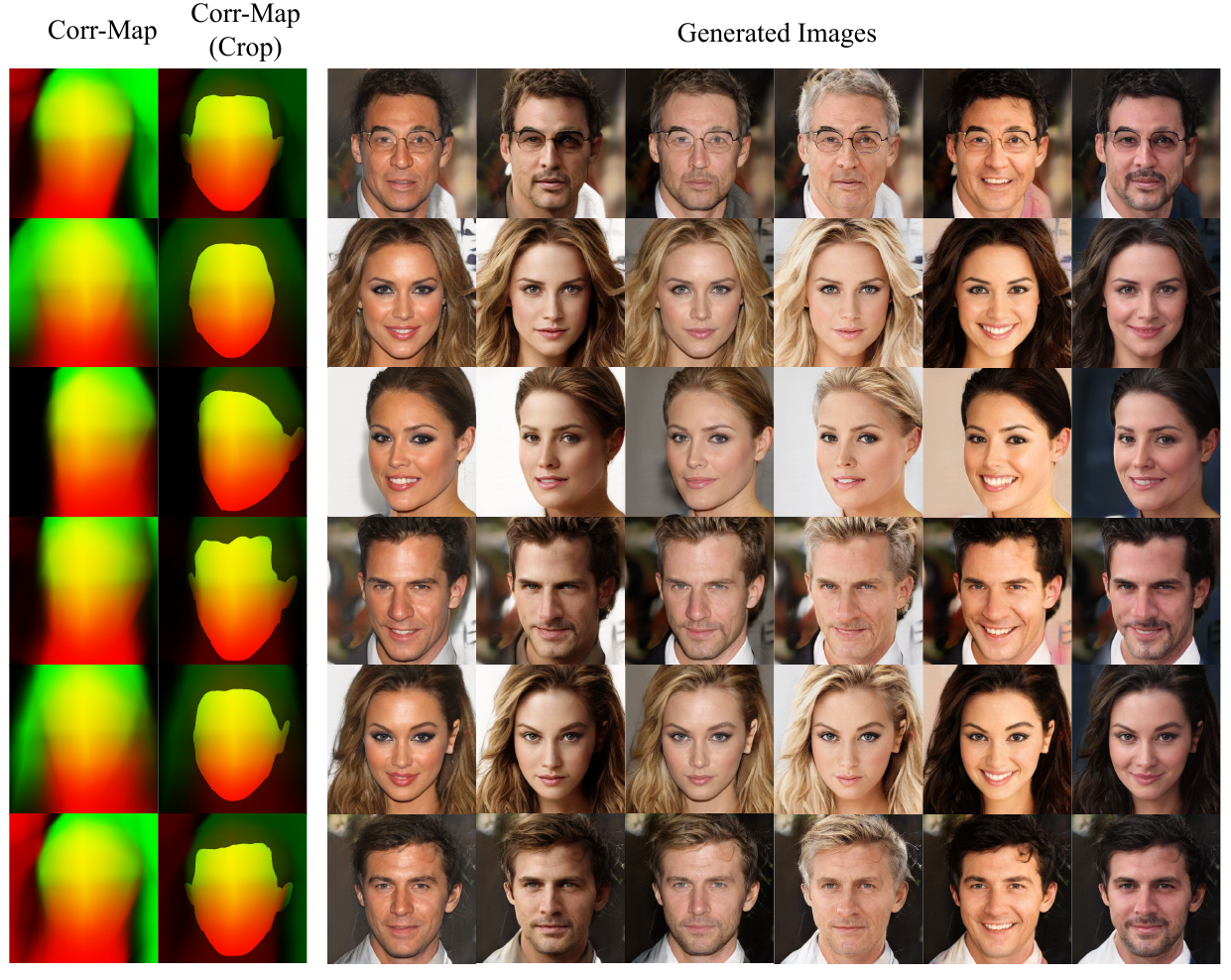}
    \vspace{-0.30in}
    \caption{Images synthesised by the proposed CoordGAN model: each row displays images with the same structure but different textures; in each column, structure varies while texture is fixed. The correspondence maps (Corr-Map) controlling the structure of the synthesized images are shown in the first column of each row. For better visualization, we use off-the-shelf segmentation models to highlight the foreground areas of all the predicted correspondence maps, as shown with Corr-Map (Crop).}
    \label{fig:face3}
    \vspace{-0.15in}
\end{figure*}

\begin{figure*}[t]
    \centering
    \includegraphics[width=\linewidth]{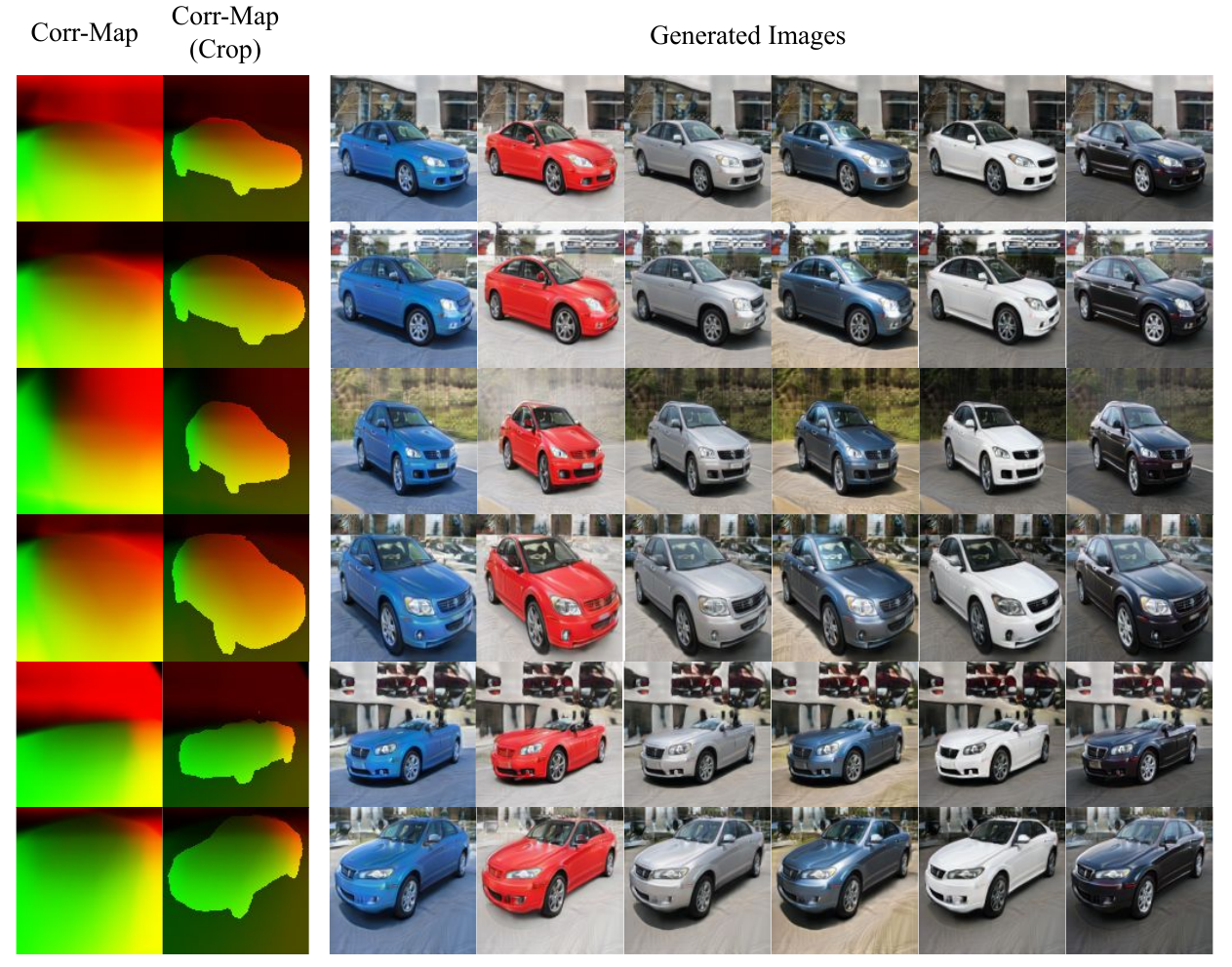}
    \vspace{-0.30in}
    \caption{Images synthesised by the proposed CoordGAN model: each row displays images with the same structure but different textures; in each column, structure varies while texture is fixed. The correspondence maps (Corr-Map) controlling the structure of the synthesized images are shown in the first column of each row. For better visualization, we use off-the-shelf segmentation models to highlight the foreground areas of all the predicted correspondence maps, as shown with Corr-Map (Crop).}
    \label{fig:car1}
    \vspace{-0.15in}
\end{figure*}

\begin{figure*}[t]
    \centering
    \includegraphics[width=\linewidth]{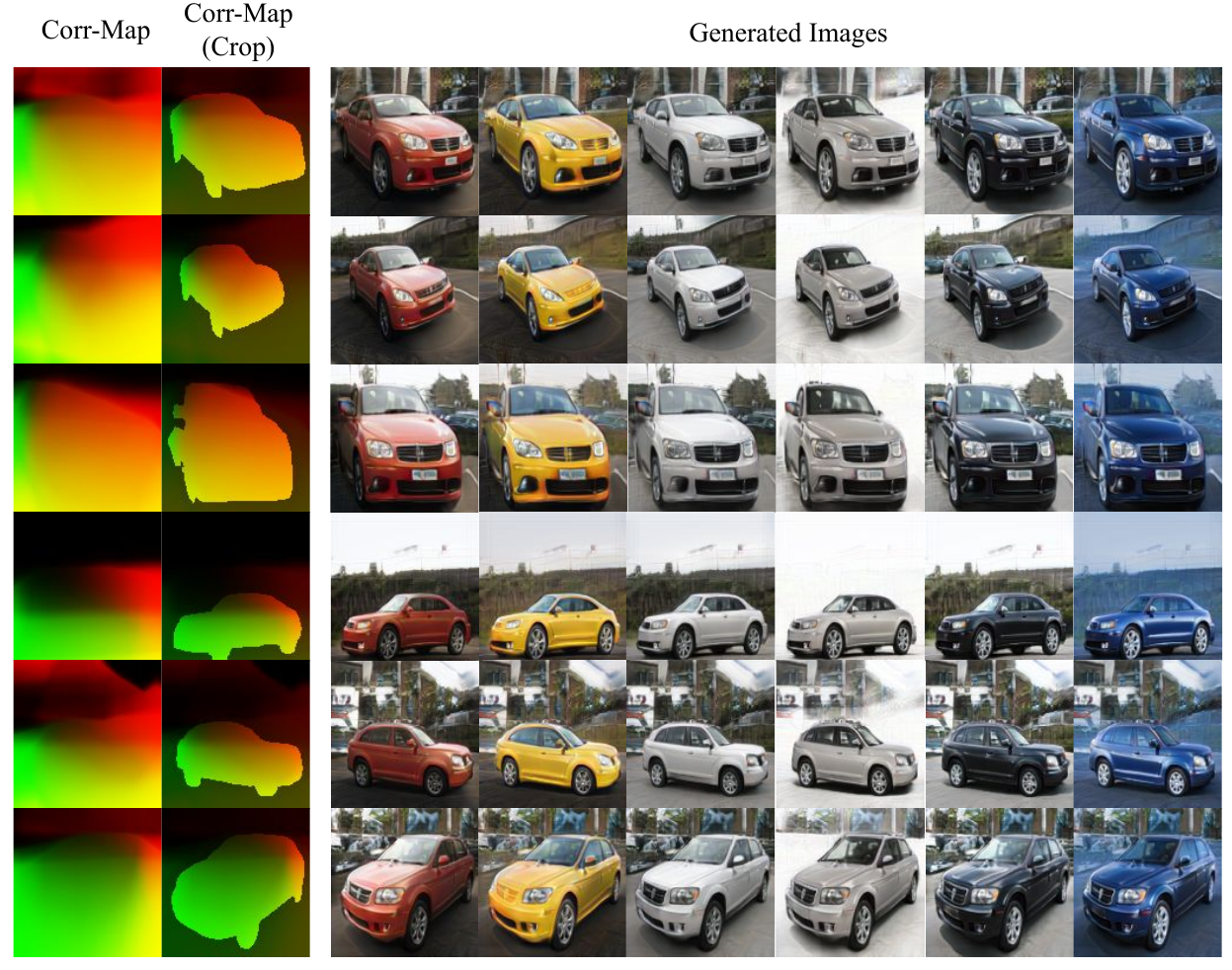}
    \vspace{-0.30in}
    \caption{Images synthesised by the proposed CoordGAN model: each row displays images with the same structure but different textures; in each column, structure varies while texture is fixed. The correspondence maps (Corr-Map) controlling the structure of the synthesized images are shown in the first column of each row. For better visualization, we use off-the-shelf segmentation models to highlight the foreground areas of all the predicted correspondence maps, as shown with Corr-Map (Crop).}
    \label{fig:car2}
    \vspace{-0.15in}
\end{figure*}

\begin{figure*}[t]
    \centering
    \includegraphics[width=\linewidth]{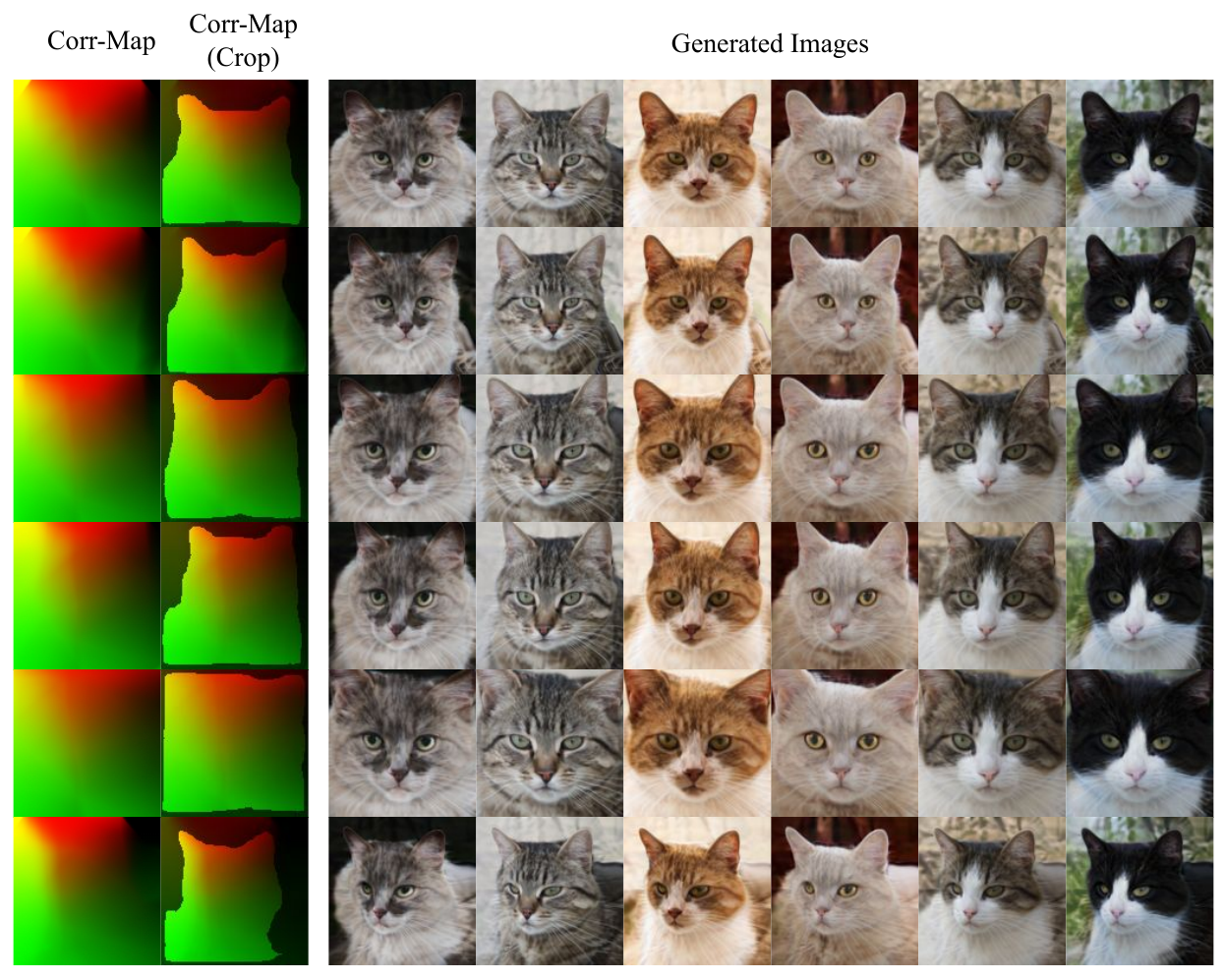}
    \vspace{-0.30in}
    \caption{Images synthesised by the proposed CoordGAN model: each row displays images with the same structure but different textures; in each column, structure varies while texture is fixed. The correspondence maps (Corr-Map) controlling the structure of the synthesized images are shown in the first column of each row. For better visualization, we use off-the-shelf segmentation models to highlight the foreground areas of all the predicted correspondence maps, as shown with Corr-Map (Crop).}
    \label{fig:cat1}
    \vspace{-0.15in}
\end{figure*}

\begin{figure*}[t]
    \centering
    \includegraphics[width=\linewidth]{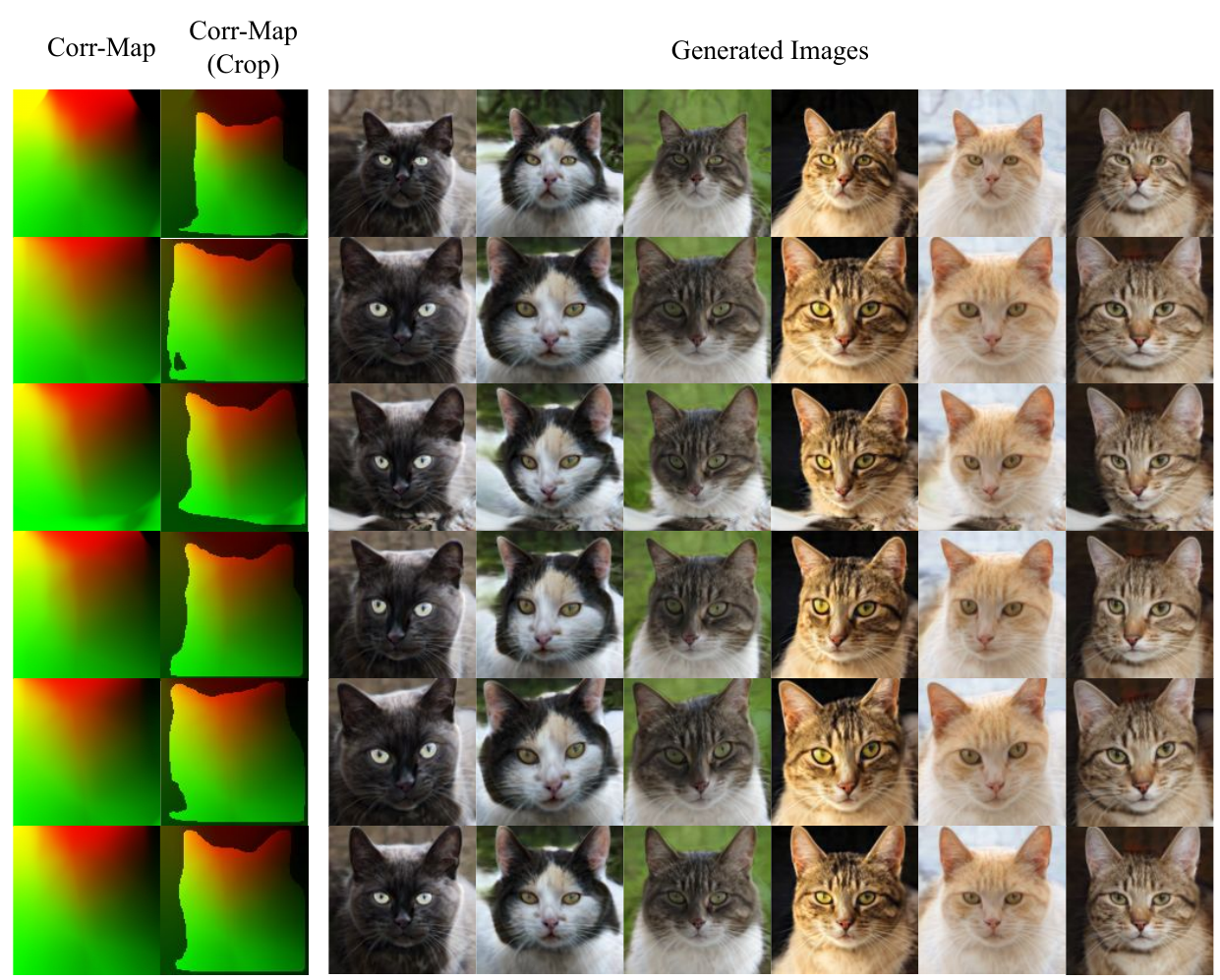}
    \vspace{-0.30in}
    \caption{Images synthesised by the proposed CoordGAN model: each row displays images with the same structure but different textures; in each column, structure varies while texture is fixed. The correspondence maps (Corr-Map) controlling the structure of the synthesized images are shown in the first column of each row. For better visualization, we use off-the-shelf segmentation models to highlight the foreground areas of all the predicted correspondence maps, as shown with Corr-Map (Crop).}
    \label{fig:cat2}
    \vspace{-0.15in}
\end{figure*}

Table~\ref{table:amt} shows that CoordGAN significantly outperforms DiagonalGAN in terms of texture-swap ratings on both CelebAMask-HQ and Stanford Cars datasets. This further suggests that the proposed approach of modeling the structure with a coordinate space effectively disentangles fine-grained structure from texture. While both methods perform similarly in terms of structure-swap studies, we emphasize that many structure-swapped pairs from DiagonalGAN are just slightly different as the learned structure code is only responsible for coarse viewpoint. More visualization results are shown in ~\Cref{fig:face1,fig:face2,fig:face3,fig:car1,fig:car2,fig:cat1,fig:cat2}.

    \begin{table}[t]
    \fontsize{8}{10pt}\selectfont
    \setlength{\tabcolsep}{2pt}
    \centering
    \begin{tabular}{l|cc|cc} 
     &\multicolumn{2}{c|}{CelebAMask-HQ} & \multicolumn{2}{c}{Stanford Cars} \\
     & Struc-swap $\uparrow$ & Text-swap $\uparrow$ & Struc-swap $\uparrow$ & Text-swap $\uparrow$ \\
    \shline
    DiagonalGAN  & 3.39 & 2.83 & 3.76 & 3.11\\
    CoordGAN & 3.32 & 3.68 & 3.58 & 3.77\\

    \end{tabular}
    \caption{User study on attribute swapping. Struc-swap denotes the setting where the pair of images are generated with the same texture code but different structure codes; Text-swap denotes the setting where the pair of images are generated with the same structure code but diverse textures.}
    \label{table:amt}
    \end{table}

\section{Coordinate Warping Network Analysis}\label{sec:backward-mlp}

\begin{figure}[t]
    \centering
    \includegraphics[width=\linewidth]{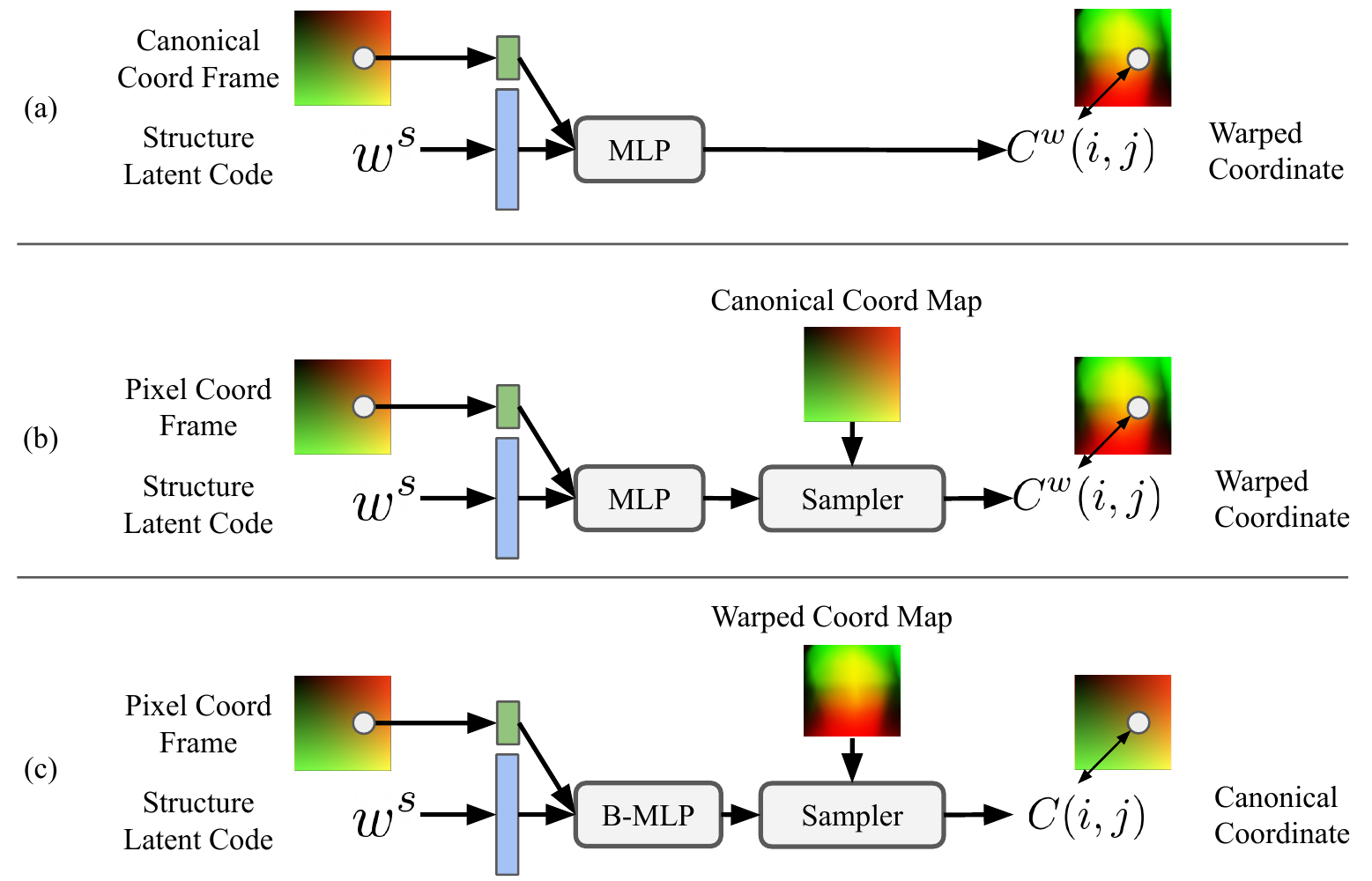}
    \caption{Coordinate warping network design. Sampler indicates the grid sampling operation.}
    \label{fig:backward-mlp}
\end{figure}

In this section, we validate that the coordinate warping network, designed as an MLP conditioned on the sampled structure code, formulates an explicit geometric transformation between the canonical coordinate frame and a warped coordinate frame. Formally, a geometric transformation between two coordinate frames should satisfy two properties: (1) one-to-one mapping exists between each element of two sets; (2) the transformation is invertible.

In the following, we show that the coordinate warping network equivalently outputs a flow w.r.t. each input coordinate, which satisfies the aforementioned property (1). We begin by defining another pixel coordinate frame $\mathcal{P}$ denoting pixel locations. This is numerically similar to the canonical coordinate frame, where coordinates are normalized to the range [-1, 1]. For example, $\mathcal{P}(1,1) = (1,1)$ indicates the bottom right pixel is of value (1,1). It then follows that the proposed coordinate warping network, as shown in Figure~\ref{fig:backward-mlp} (a), is equivalent to the architecture in Figure~\ref{fig:backward-mlp} (b). This comes from two facts: (i) the pixel coordinate frame is constructed exactly the same as the canonical coordinate frame; (ii) the grid sampling operation in Figure~\ref{fig:backward-mlp} (b) outputs exactly the same value as the MLP output as the MLP is constrained to output values from -1 to 1. Therefore, we show that, given a structure code, the MLP learns a transformation from the canonical coordinate frame to the warped coordinate frame.

In addition, we build another backward MLP to satisfy the second property, such that warped coordinates can be back to canonical coordinates with the same structure code. Specifically, as shown in Figure~\ref{fig:backward-mlp} (c), we construct another three-layer MLP to map warped coordinates to canonical coordinates. To distinguish the MLP mapping from canonical coordinates to warped coordinates, we refer to this one as the Backward MLP (B-MLP). We train CoordGAN with the additional B-MLP (CoordGAN-B) from scratch, where the B-MLP is supervised with an additional L1 loss between the predicted canonical coordinate frame and ground-truth canonical coordinate frame. As show in Table~\ref{table:backward-mlp generation} and Table~\ref{table:backward-mlp correspondence}, CoordGAN-B achieves on average better performance in both image synthesis and label propagation. 

To this end, we prove that the proposed MLP models an explicit geometric transformation between the canonical coordinate frame and a warped coordinate frame. We opt for an MLP as it preserves the order of the coordinates in the canonical coordinate frame due to its continuity. Since an explicit transformation is learned, it ensures that, when the MLP outputs the same coordinate given two different structure codes, these two positions are corresponding to the same coordinate in the canonical frame.

    \begin{table}[t]
    \fontsize{8}{12pt}\selectfont
    \setlength{\tabcolsep}{4pt}
    \centering
    \begin{tabular}{l|ccc|cc} 
     &\multicolumn{3}{c|}{CelebA-HQ} & \multicolumn{2}{c}{Stanford Cars} \\
     & LPIPS $\downarrow$ & Arcface $\downarrow$ & FID $\downarrow$ & LPIPS $\downarrow$ & FID $\downarrow$\\
    \shline
    CoordGAN & 0.22 & 0.38 & 16.16 & 0.21 & 24.27 \\
    CoordGAN-B & 0.19 & 0.38 & 15.45 & 0.18 & 23.95 \\
    \end{tabular}
        \vspace{-0.10in}
    \caption{Image generation results on Coordinate Warping Network with backward MLP (CoordGAN-B).}
        \vspace{-0.1in}
    \label{table:backward-mlp generation}
    \end{table}
    
    \begin{table}[t]
    \fontsize{8}{12pt}\selectfont
    \setlength{\tabcolsep}{4pt}
    \centering
    \begin{tabular}{l|ccc}
     & CelebA-HQ& DGAN-face & DGAN-car \\
    \shline
    CoordGAN & 52.25 & 23.78 & 13.23  \\
    CoordGAN-B & 54.51 & 25.44 & 12.59  \\
    \end{tabular}
        \vspace{-0.10in}
    \caption{Label propagation results on Coordinate Warping Network with backward MLP (CoordGAN-B). Measured by IOU.}
        \vspace{-0.1in}
    \label{table:backward-mlp correspondence}
    \end{table}

\section{Implementation Details}\label{sec:imp}
We introduce the training details and specify the architecture for each module of our network.

\subsection{Architecture}

\textbf{Generator.}
Both the sampled structure and texture codes are 512-dimensional. The structure and texture mapping networks are implemented with an 8-layer MLP with a latent dimension of 512. The coordinate warping network, conditioned on a latent structure code, is implemented with a three-layer MLP. A tanh function is used at the output of the coordinate mapping network to ensure that the output is within a valid coordinate space. The dense correspondence map is passed to a positional encoding layer where, a Fourier embedding with 512 channels is obtained by the application of a $1 \times 1$ convolution followed by a sine function. In all experiments, the canonical coordinate map and the correspondence map are defined with a spatial resolution of 128. The modulated generator consists of 10 layers and all layers are with 512 channels. The design of each layer is similar to StyleGAN2. We follow StyleGAN2 to inject the latent texture code into different layers of the modulated generator via weight modulation/demodulation. The dense correspondence map is concatenated with all 10 layers of the modulated generator, as shown in Figure~\ref{fig:encoder}. To generate higher resolution images, another two upsampling blocks are added to the last layer of the modulated generator. Note that the correspondence map is not concatenated to these upsampling blocks. Skip connections are used to combine features for every two layers from intermediate feature maps to RGB values.

\textbf{Patch Discriminator.}
The patch discriminator architecture for the structure-swapping constraint is designed following Swapping Autoencoder. The patch discriminator consists of a feature extractor of 5 downsampling residual blocks, 1 residual block, and 1 convolutional layer, and a classifier. Specifically, 8 randomly cropped patches from the same image are used as reference. Each patch is cropped randomly from $\frac{1}{8}$ to $\frac{1}{4}$ of the image dimensions for each side. All cropped patches are resized to $\frac{1}{4}$ of the image size and then input to the patch discriminator. Each patch is passed to multiple downsampling blocks to obtain a feature vector. The feature vectors of all reference patches are averaged and then concatenated with a feature vector from a real or fake patch. The real patches are patches from the same image as the reference patches and fake patches are from a structure-swapping image. The classifier finally determines whether the concatenated feature vector is real or fake.

\textbf{Encoder.}
Given an image, the encoder produces two 512 dimensional vectors. As shown in Figure~\ref{fig:encoder}, our encoder network design follows Swapping Autoencoder. The difference is that instead of outputting a feature map for the structure code, the proposed design outputs a 512 dimensional structure code. Specifically, 4 downsampling residual blocks are first applied to produce an intermediate tensor, which then produces separate features for the structure code and texture codes. The structure code is produced by first applying 1-by-1 convolutions to the intermediate tensor, reducing the number of channels and then applying a fully-connected layer. The texture codes in the W+ space are produced by applying stride convolutions, average pooling, and then different dense layers.

\subsection{Training Details}

To train the generator, we follow StyleGAN2 and use the non-saturating GAN loss and lazy R1 regularization. The R1 regularization is also applied to the patch discriminator. The weight of the R1 regularization is 10.0 for the image discriminator and is 1.0 for the patch discriminator. We use the ADAM optimizer with a learning rate of 0.002 and with $\beta_1 = 0.0 $ and $\beta_2 = 0.99$. The batch size is set to 16 with 8 GPUs. Coefficients for different losses are set as following: $\lambda_{cham}=100$, $\lambda_{GAN}=2$, $\lambda_{t}=5$, $\lambda_{warp}=5$, $\lambda_{s}=1$. To warm up training, for the first 20k iterations, $\lambda_{warp}$, $\lambda_{t}$, and $\lambda_{s}$ are linearly increased from 0.  For celebAMask-HQ, we train the generator for 300k iterations at the resolution $128 \times 128$ and then train at a high resolution for another 200k iterations. For the Stanford Cars and AFHQ-cat datasets, we train the generator for 300k iterations at the resolution $128 \times 128$. The hyper parameters for training the encoder are selected as following: $\lambda_{rec}=10$, $\lambda_{con}=10$, $\lambda_{t}=5$. The encoder is trained for 200k iterations.

\section{Application in Other domains}\label{sec:metface}

\begin{figure*}[t]
    \centering
    \includegraphics[width=\linewidth]{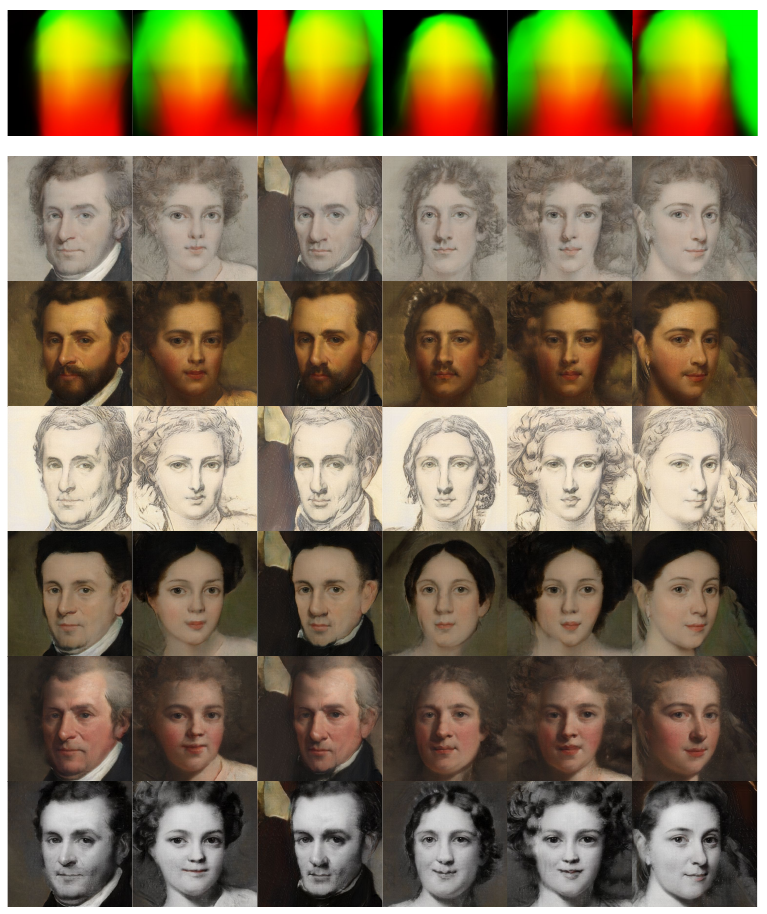}
    \vspace{-0.30in}
    \caption{Images synthesised by the proposed CoordGAN model: the first row displays correspondence maps; from the second row to the bottom, each row displays images with the same texture but different structures; in each column, texture varies while structure is fixed.}
    \label{fig:metface}
    \vspace{-0.15in}
\end{figure*}

In this section, we show that the CoordGAN can handle structure texture transfer on other domains, e.g., paintings. Specifically, we finetune the CelebAMask-HQ pre-trained model at the resolution of $512 \times 512$ on the metfaces dataset~\cite{karras2020training}. The metfaces dataset contains 1336 high-quality images at $1024 \times 1024$ resolution. Following~\cite{mo2020freeze}, we freeze the first three high resolution layers of the discriminator during finetuning. Furthermore, to enable texture swapping across different domains, we fix the weights of the structure mapping network and coordinate mapping network. As show in Figure~\ref{fig:metface}, we qualitatively demonstrate that,  CoordGAN can generate arts with high quality by combining the structure representation learned from real images with texture codes learned from arts. Note that the structure-texture disentanglement is still well maintained.

\end{document}